\documentclass[sigplan, 10pt, twocolumn]{acmart}
\renewcommand\footnotetextcopyrightpermission[1]{}
\usepackage{pifont}
\usepackage{balance}
\usepackage{enumitem}
\usepackage{subfigure}
\usepackage{multirow}

\usepackage{tablefootnote}
\usepackage{threeparttable}

\usepackage[ruled, vlined, linesnumbered, ruled]{algorithm2e}

\usepackage{color}
\usepackage[noend]{algpseudocode}
\usepackage{xspace}
\usepackage{fancyhdr}
\usepackage{multicol}
\usepackage{amsfonts}
\usepackage{ulem}

\usepackage{tikz}
\newcommand*{\circled}[1]{\lower.8ex\hbox{\tikz\draw (0pt, 0pt)%
    circle (.47em) node {\makebox[0.4em][c]{\small #1}};}}

\def\ie{\textit{i.e.}\xspace}

\def\eg{\textit{e.g.}\xspace}

\def\method{\textsf{Caesar}\xspace}
\def\fedavg{\textsf{FedAvg}\xspace}
\def\flexcom{\textsf{FlexCom}\xspace}
\def\prowd{\textsf{ProWD}\xspace}
\def\pyramidfl{\textsf{PyramidFL}\xspace}
\newcommand{\tabincell}[2]{\begin{tabular}{@{}#1@{}}#2\end{tabular}}

\AtBeginDocument{%
  }

\setcopyright{acmcopyright}
\copyrightyear{2025}
\acmYear{2025}
\acmDOI{xxxxxx}

\begin{document}
 
\title{Caesar: Efficient Federated Learning via Low-deviation Model and Gradient Compression}

\author{Jiaming Yan$^{1, 2}$, Jianchun Liu$^{1, 2}$, Hongli Xu$^{1, 2}$, Liusheng Huang$^{1, 2}$, \\
Jiantao Gong$^{3}$, Xudong Liu$^{3}$, Kun Hou$^{3}$}

\affiliation{
    \institution{$^{1}$ School of Computer Science and Technology, University of Science and Technology of China}
    \country{}
}

\affiliation{
    \institution{$^{2}$ Suzhou Institute for Advanced Research, University of Science and Technology of China}
    \country{}
}

\affiliation{
    \institution{$^{3}$ Guangdong OPPO Mobile Telecommunications Corp., Ltd.}
    \country{}
}

\email{jmyan@mail.ustc.edu.cn, {jcliu17, xuhongli, lshuang}@ustc.edu.cn, {gongjiantao, liuxudong, houkun}@oppo.com}

\fancyhf{} 

\begin{abstract}
Compression is an efficient way to relieve the tremendous communication overhead of federated learning (FL) systems.
However, for the existing works, the information loss under compression will lead to unexpected model/gradient deviation for the FL training, significantly degrading the training performance, especially under the challenges of data heterogeneity and model obsolescence.
To strike a delicate trade-off between model accuracy and traffic cost, we propose \method, a novel FL framework with a low-deviation compression approach.
For the global model download, we design a greedy method to optimize the compression ratio for each device based on the staleness of the local model, ensuring a precise initial model for local training.
Regarding the local gradient upload, we utilize the device's local data properties (\ie, sample volume and label distribution) to quantify its local gradient's importance, which then guides the determination of the gradient compression ratio.
Besides, with the fine-grained batch size optimization, \method can significantly diminish the devices' idle waiting time under the synchronized barrier.
We have implemented \method on two physical platforms with 40 smartphones and 80 NVIDIA Jetson devices.
Extensive results show that \method can reduce the traffic costs by about 25.54\%$\thicksim$37.88\% compared to the compression-based baselines with the same target accuracy, while incurring only a 0.68\% degradation in final test accuracy relative to the full-precision communication.

\end{abstract}



\keywords{Federated Learning, Model Obsolescence, Data Heterogeneity, Deviation-aware Compression.}

\maketitle

\pagenumbering{arabic}
\fancyfoot[C]{\fontsize{10pt}{40pt}\selectfont\thepage}
\setcounter{page}{1}
\thispagestyle{fancy}
\section{Introduction}\label{sec_introduction}
The rapid growth of data generated by mobile devices has driven AI applications like e-commerce recommendations \cite{niu2020billion} and autonomous driving \cite{zheng2023autofed}, but also raises privacy concerns, leading to regulations such as GDPR \cite{enwiki:1208869058} and CCPA \cite{enwiki:1198023421}.
To this end, Google proposed federated learning (FL) \cite{mcmahan2017communication} to train the AI models in a privacy-preserving manner.
In practice, the FL systems typically need to involve large-scale (hundreds or even thousands) devices and spend enormous communication rounds to obtain high-accuracy models \cite{liu2023yoga, liu2023adaptive}.
With the increasing size of AI models, the massive transmission of models and gradients between devices and the PS will incur significant communication overhead \cite{liu2021adaptive, jiang2022fedmp}.
For example, training BERT-base \cite{kenton2019bert} on 20news \cite{lang1995newsweeder} via FL will occur a total traffic volume of up to 90.15TB \cite{lin2021fednlp}.
Such huge transmission payloads will result in long communication latency and expensive traffic charges for manufacturers (\eg, \$0.01/GB on AWS \cite{cai2023efficient}), contradicting their interests.
Therefore, \textit{how to achieve the trade-off between model accuracy and traffic cost} becomes a critical issue for FL \cite{nguyen2021federated, kairouz2021advances, jiang2023heterogeneity, wang2023accelerating}.

However, there are two key challenges that bottleneck FL in achieving the above trade-off.
\textbf{\textit{1) Data Heterogeneity.}}
Since each device individually collects its own data, which is related to geographical location and/or user preferences, the data distribution would vary significantly across devices, \ie, non-independent and identically distributed (non-IID) \cite{wang2020optimizing, liu2022enhancing}, so as the data volume.
Such data heterogeneity will affect the training performance of FL, decelerating the convergence rate and potentially compromising the final model accuracy.
Therefore, with the data heterogeneity, more communication rounds (\ie, more traffic costs) are required to achieve the target accuracy in the FL systems.
\textbf{\textit{2) Model Obsolescence.}}
In practice, the PS selects only a small fraction of devices to participate in each communication round \cite{bonawitz2019towards}.
Some devices may not be selected for many rounds, thereby they will retain obsolete local models that significantly differ from the current global model. 
Besides, since devices might be unavailable for selection at times (\eg, loss of connection, out of battery), the participation frequency of different devices may be various.
As a result, different devices usually hold the local models with diverse staleness (\ie, the difference with the latest global model).

\textbf{Status Quo and their Limitations.}
Compressing the transmission payload, including global model and local gradient, is the most straightforward and efficacious way to reduce traffic costs, prompting the development of numerous compression-based FL schemes \cite{cui2022optimal, xu2021deepreduce, dorfman2023docofl, bernstein2018signsgd, li2021talk, xu2022adaptive, mei2022resource, liu2023communication}. The previous works can be broadly classified into two categories based on their different strategies to configure the model/gradient compression ratio.
The first category of schemes \cite{cui2022optimal, xu2021deepreduce, dorfman2023docofl, bernstein2018signsgd} simply adopts a fixed and identical compression ratio for all devices.
However, devices often have heterogeneous capabilities (\eg, bandwidth, computing power), resulting in varying round completion times. 
Powerful clients should wait for weak ones for global aggregation, and such idle waiting will significantly depress the training efficiency of FL \cite{yan2024peaches}.
To tackle this issue, the second category of schemes \cite{li2021talk, xu2022adaptive, mei2022resource, liu2023communication} adjusts the compression ratio according to the devices' capabilities.
Generally, weaker devices tend to use larger compression ratios to shorten the completion time and mitigate the idle waiting for stronger devices.
Nevertheless, these capability-aware schemes would hurt the training performance of FL, especially in the data heterogeneity scenarios.
For example, devices with high-quality local data (\eg, balanced label distribution and/or large sample volume) will derive the gradients that are crucial for model convergence \cite{ye2023feddisco}.
Yet, if these devices have poor capabilities, their important gradients can be easily corrupted due to the excessive compression, leading to low model accuracy.

Besides, the challenge of model obsolescence further diminishes the performance of both compression schemes.
Specifically, if a large compression ratio is adopted for model download, devices with obsolescent local models may struggle to accurately recover the compressed global model.
As a result, these devices will conduct local training with an imprecise initial model, producing gradients that are ineffective or even detrimental to global updates.
On the contrary, utilizing a small model compression ratio may cause unnecessary traffic costs on devices with fresh local models.
This is because these devices can construct an accurate initial model for training even with an over-compressed global model. 
In summary, while compression strategies are essential for reducing communication overhead in FL, the existing schemes will significantly hurt the training performance under the challenges of data heterogeneity and model obsolescence.

\textbf{Overview of the Proposed Framework.} 
To fill this critical gap, we propose \method, a novel FL framework, with a low-deviation compression approach.
At the server side, \method cherry-picks the global model compression ratio for devices based on their local models' staleness, which allows every participant to obtain a precise initial model for local training.
At the device side, the local gradient compression is guided by the local data properties.
For instance, the devices with high-quality data will adopt small compression ratios to reduce the deviation (\ie, compression error) on critical gradients.
Although our proposed compression approach is expected to solve the two critical challenges, the training efficiency still suffers from the straggler problem.
For example, devices using low model and/or gradient compression ratios may experience high communication latency if their bandwidth is limited, leading to non-negligible waiting time for other participants.
To address this issue, \method optimizes batch size configurations for different devices in a fine-grained manner, improving training efficiency without affecting model accuracy.

Nevertheless, there is a thorny difficulty for the system design of \method.
In real-world networks, it is challenging to determine the optimal model/gradient compression ratios for diverse devices.
On the one hand, since devices participate in training with different frequencies, the staleness of local models can be various.
On the other hand, the local data properties of devices are also heterogeneous, resulting in local gradients with varying importance.
There is no silver-bullet configuration for model and gradient compression ratios that achieves the trade-off between model accuracy and traffic cost for all devices.
Thus, \textit{how to determine the appropriate global model and local gradient compression ratios for heterogeneous devices} is the core challenge for the design of \method.
The main contributions of this paper can be summarized as follows:
\begin{itemize}
    \item We design a novel FL framework, termed \method, to achieve a delicate trade-off between model accuracy and traffic cost under the challenges of data heterogeneity and model obsolescence.
    \item For the global model compression, we establish a relationship between the device's local staleness and model compression ratio based on the greedy strategy.
    Then, we further optimize the process of model compression to mitigate the information loss.
    \item For the local gradient compression, we formulate the importance of each device's gradient on the basis of its data properties and propose a rank-based algorithm to determine the proper gradient compression ratio.
    \item We implement \method on two physical platforms with 40 OPPO smartphones and 80 NVIDIA Jetson devices. 
    Extensive results show \method's superior traffic-to-accuracy performance over the baselines.
\end{itemize}

\section{Background and Motivation}\label{sec:prelim}
\begin{figure*}[t]
	\centering
        \subfigure[Training Performance]{
		\includegraphics[width=1.58in]{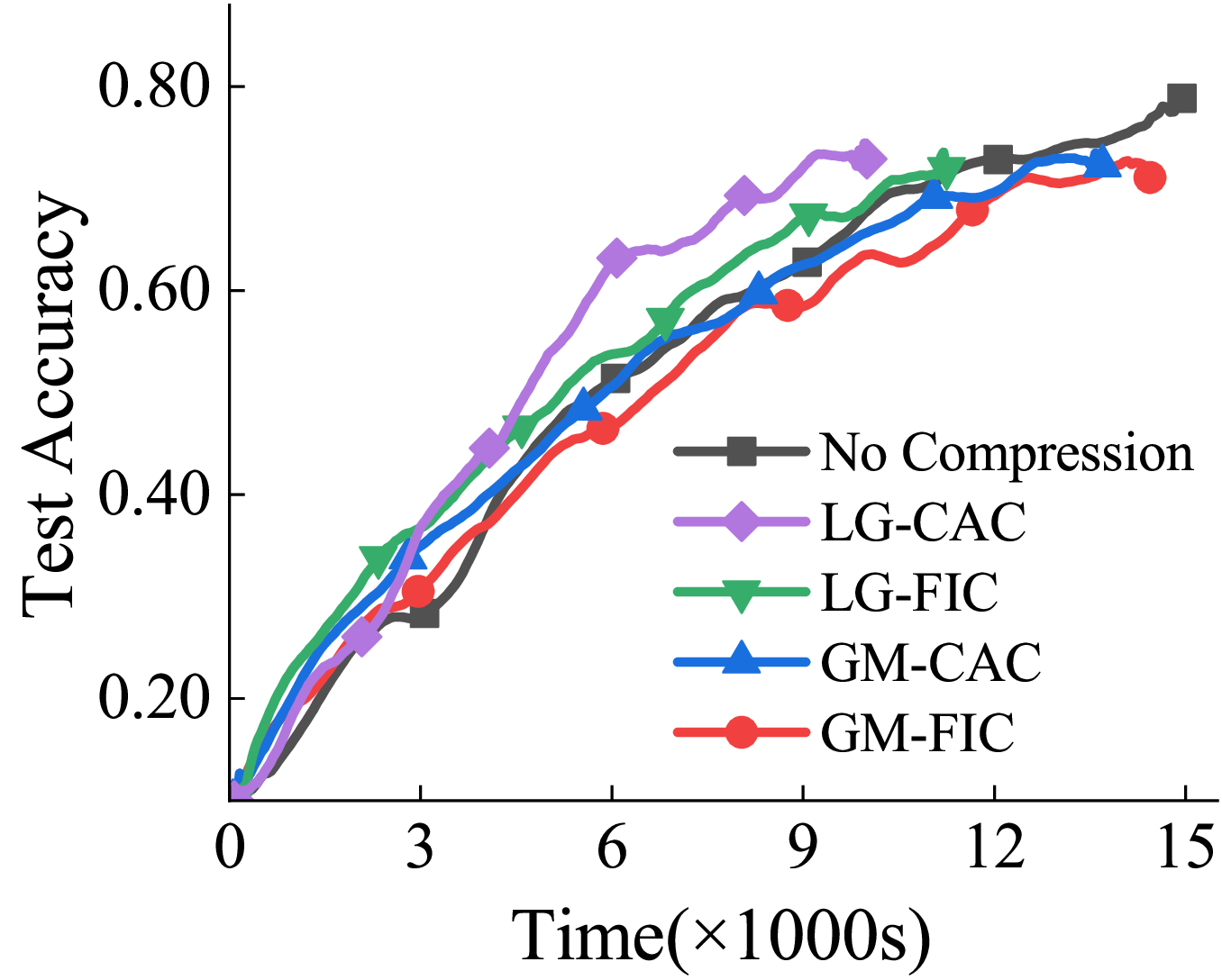}\label{fig:motivation_a}
	}
	\subfigure[Traffic Cost]{
		\includegraphics[width=1.58in]{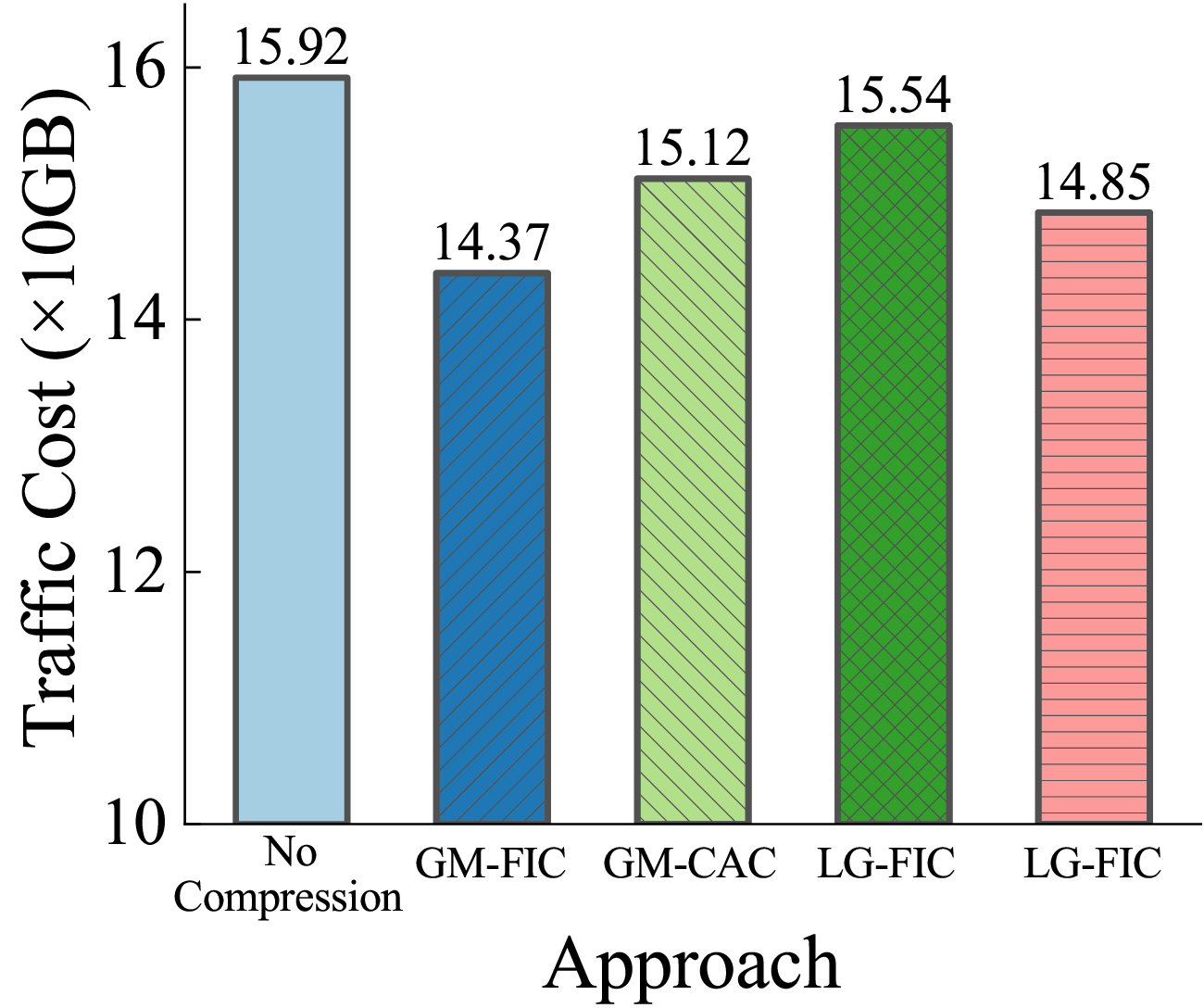}\label{fig:motivation_b}
	}
 \subfigure[Initial Model Error]{
		\includegraphics[width=1.58in]{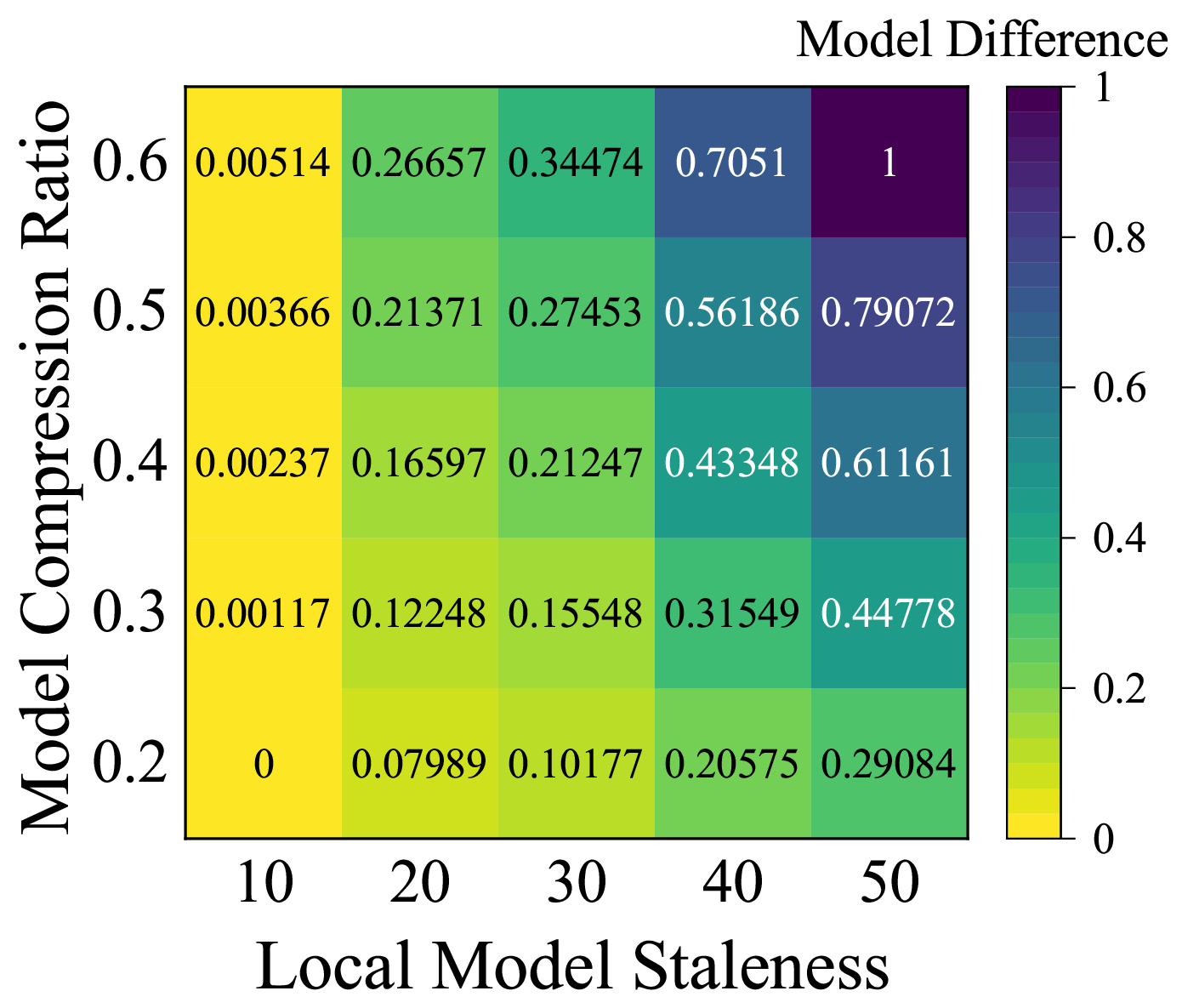}\label{fig:motivation_c}
	}
        \subfigure[Local Gradient Importance]{
		\includegraphics[width=1.58in]{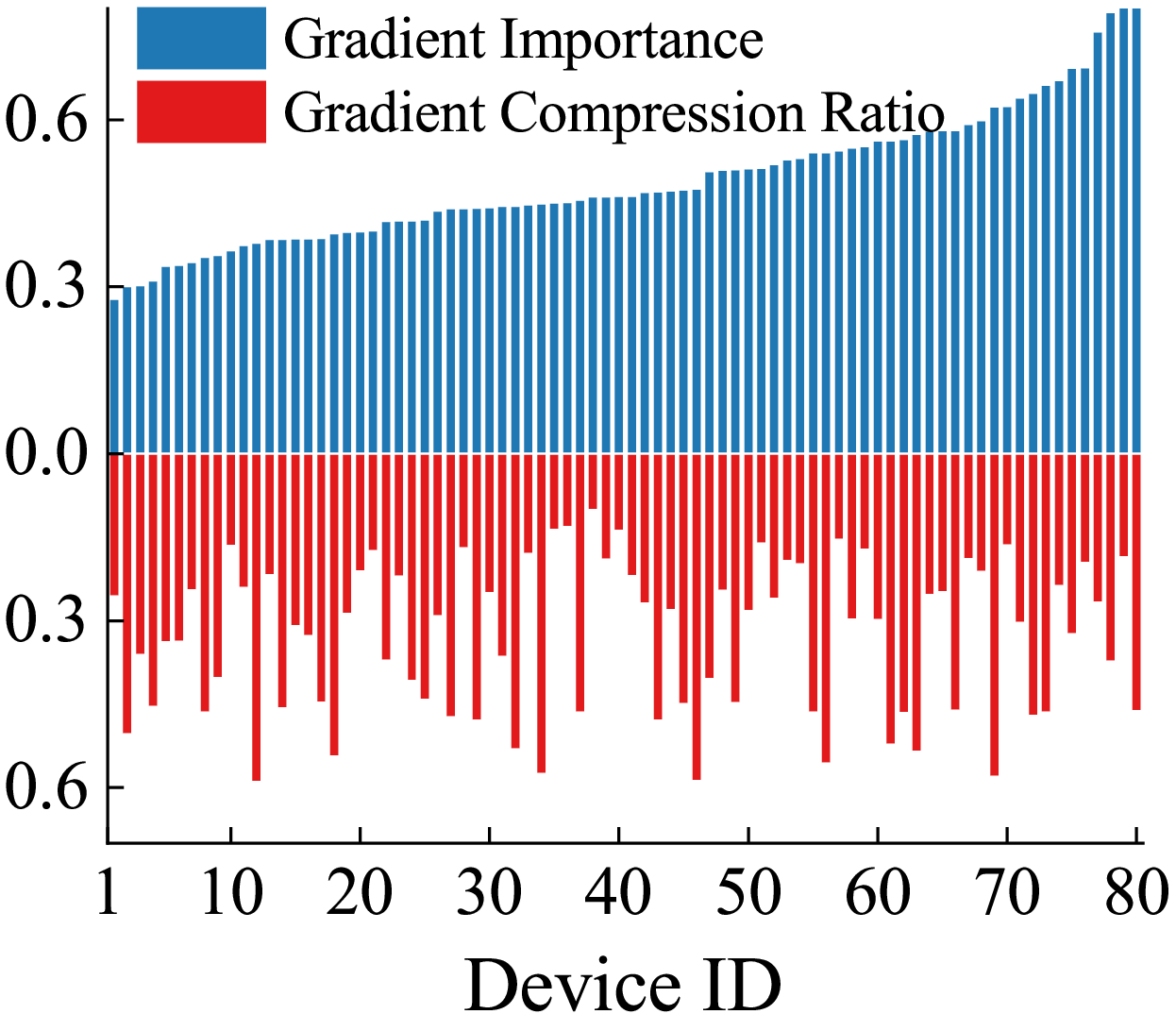}\label{fig:motivation_d}
	}
 \vspace{-3mm}
	\caption{The results of preliminary experiments. 
(a) The training process of different FL approaches on CIFAR-10 with 250 communication rounds.
(b) The traffic costs of different FL approaches to achieve a target accuracy of 72\% on CIFAR-10;
(c) The relationship between initial model error, model compression ratio, and local model staleness;
(d) The importance of devices' local gradients and the adopted gradient compression ratio with \textsf{CAC}.
 }\label{fig:motivation}
\end{figure*}

\subsection{FL with Compressed Communication}
In a standard FL system, there is a set of $n$ devices $\mathbf{N} = \{v_{1}, v_{2}, \cdots, v_{n} \}$ coordinated by a PS.
Each device $v_{i} \in \mathbf{N}$ hosts the local dataset $\mathcal{D}_{i}$ with the data volume $| \mathcal{D}_{i} | = m_{i}$.
Let $l(w; \xi)$ denote the loss function, where $w$ is the model and $\xi$ represents the data sample.
The optimization objective of FL can be formulated as \cite{wang2023bose}:
\begin{equation}
    \min_{w} F(w) := \min_{w} \sum_{i = 1}^{n} \frac{m_{i}}{m} \mathbb{E}_{\xi \thicksim \mathcal{D}_{i}} l(w; \xi)
\end{equation}
where $m = \sum_{i = 1}^{n} m_{i}$ is the total data volume.
Suppose that a total of $T$ communication rounds are required until convergence. 
In each round $t \in [T]$, the PS selects a subset of devices $\mathbf{N}^{t}$ for model training with a participation fraction $\alpha \in (0, 1]$.
Then, the latest global model $w^{t}$ will be compressed and sent to participants.
Each selected device $v_{i} \in \mathbf{N}^{t}$ receives the compressed global model $\overline{w}_{i}^{t}$ and tries to recover it with its local model $w_{i}$.
After that, the recovered model $\widehat{w}_{i}^{t}$ will be trained on the local data for $\tau$ iterations via the mini-batch SGD optimizer \cite{li2014efficient} in each round $t$ as:
\begin{equation}
    w_{i}^{t, j + 1} \leftarrow w_{i}^{t, j} - \eta^{t} \nabla l(w_{i}^{t,j}; \zeta_{i}^{t, j}), \quad j \in \{ 0, 1, \cdots, \tau - 1 \}
\end{equation}
where $w_{i}^{t, j}$ is the local model of device $v_{i}$ in round $t$ after $j$ iterations, and $w_{i}^{t, 0} = \widehat{w}_{i}^{t}$.
$\eta^{t}$ and $\zeta_{i}^{t, j}$ represent the learning rate and a random data batch sampled from the local dataset $\mathcal{D}_{i}$ in iteration $j$ at round $t$, respectively.
After the local training, each device $v_{i}$ renews its local model, \ie, $w_{i} \leftarrow w_{i}^{t, \tau}$, and derives the local gradient $g_{i}^{t} = \eta^{t} \sum_{j = 0}^{\tau - 1} \nabla l(w_{i}^{t, j}; \zeta_{i}^{t, j})$.
Next, the compressed local gradient $\overline{g}_{i}^{t}$ will be uploaded to the PS for global aggregation, and then the global model will be updated, \ie, $w^{t + 1} \leftarrow w^{t} - \sum_{v_{i} \in \mathbf{N}^{t}} \frac{1}{\| \mathbf{N}^{t} \|} \overline{g}_{i}^{t}$.

\subsection{Limitation of Existing Approaches} \label{sec:motivation}
While compression is an efficient way to relieve the substantial traffic costs of FL, the determination of compression ratio is critical to both the training performance and communication efficiency \cite{jiang2023heterogeneity, wang2023accelerating}.
There are two main categories of approaches for determining the compression ratio in existing studies. 
The first category of approaches \cite{cui2022optimal, xu2021deepreduce, dorfman2023docofl, bernstein2018signsgd} simply adopts a fixed and identical compression (\textsf{FIC}) ratio for all devices.
The second category of approaches \cite{li2021talk, xu2022adaptive, mei2022resource, liu2023communication} determines the compression ratio according to the capabilities of devices, represented as capability-aware compression (\textsf{CAC}).
In simple terms, devices with small bandwidth and/or weak computational power tend to adopt large compression ratios to shorten their round duration, thereby reducing idle waiting times for stronger devices.
However, with the challenges of data heterogeneity and model obsolescence, the compression ratios determined by these approaches may be sub-optimal in achieving the best trade-off between model accuracy and traffic cost.

To illustrate the limitation of existing approaches, we conduct a set of preliminary experiments on a physical platform formed by 80 NVIDIA Jetson devices.
Specifically, we allocate the CIFAR-10 dataset to these devices via the Dirichlet distributions \cite{hsu2019measuring, yurochkin2019bayesian} and implement the training of ResNet-18 \cite{he2016deep} on this platform.
For \textsf{CAC}, we follow \cite{li2022pyramidfl} to span the compression ratio from 0.1 to 0.6 based on the capabilities of participants, while the compression ratio in \textsf{FIC} is set to 0.35.
Top-K sparsification [26] is employed for both the model and gradient compression.
More detailed descriptions of experimental setup can be found in Section \ref{sec:exp_set}.

In our preliminary experiments, four baselines are adopted for the performance comparison with the typical FL training (no model and gradient compression).
Concretely, \textsf{GM-FIC} and \textsf{GM-CAC} separately utilize \textsf{FIC} and \textsf{CAC} to compress the global model without the local gradient compression.
\textsf{LG-FIG} and \textsf{LG-CAC} compress the local gradient by \textsf{FIC} and \textsf{CAC}, respectively, without the global model compression.
As shown in Figure \ref{fig:motivation_a}, the baselines can speedup the FL training process by about 1.09$\times$$\thicksim$1.47$\times$.
This is attributed to the communication latency reduction by compression.
Since \textsf{CAC} can further reduce the waiting time among devices by considering their different capabilities, the acceleration effect of \textsf{CAC}-based approaches is more pronounced compared to \textsf{FIC}-based approaches.
However, the compression will also decrease the training performance significantly, where the final model accuracy of the baselines is 4.42\%$\thicksim$6.75\% lower than that of typical FL (\ie, No Compression). 
With such an accuracy decline, these approaches require more communication rounds to achieve the same model accuracy with the typical FL.
Therefore, although the model/gradient compression can mitigate the transmission volume within each communication round, their improvement in traffic-to-accuracy performance may be limited due to the slow convergence rate.
Specifically, by Figure \ref{fig:motivation_b}, given a target accuracy of 72\% (\ie, the highest one that can be achieved by all approaches), the baselines can only reduce the traffic cost by about 2.39\%$\thicksim$9.74\% compared to the typical FL.
In a word, by adopting the existing approaches for the model/gradient compression, the training performance of FL will degrade significantly, leading to inefficient traffic saving. 
Next, we will analyze the reasons for such limitations in detail.
 
\textbf{Limitation in Global Model Compression.}
From the perspective of global model compression, the main reason for the performance degradation is that the local training is conducted with an erroneous initial model, whose fidelity is related to the previous local model and the compressed global model.
Thus, the local model staleness (\ie, the degree of model obsolescence) and the model compression ratio will impose significant effects on the training performance.
Specifically, the staleness of each device's local model is defined as the number of rounds since its last participation until now.
We utilize mean-square error (MSE) to measure the difference between the initial model and the latest global model.
Then, we normalize the model difference to the range of $[0, 1]$.
As shown in Figure \ref{fig:motivation_c}, the difference will intensify rapidly with the increasing staleness and compression ratio.
However, neither \textsf{FIC} nor \textsf{CAC} takes this circumstance into consideration, thus the fidelity of initial model cannot be guaranteed.
For example, in \textsf{CAC}, the devices with weak capabilities are likely to adopt a large compression ratio, which will easily lead to low-fidelity initial models if their local models are obsolescent.

\textbf{Limitation in Local Gradient Compression.}
From the perspective of local gradient compression, the performance decline is attributed to the improper gradient compression error.
In the data heterogeneity scenario, the devices with uniform data distribution and/or abundant data samples have a crucial impact on the global update \cite{jiang2023heterogeneity, lai2021oort}.
However, their local gradients may be compressed too much in the prior approaches \cite{li2021talk, xu2022adaptive, mei2022resource, liu2023communication}.
We estimate the importance of each device's local gradient based on its data properties via Eq. \eqref{eq:con} in Section \ref{sec:gradient_compression_component}.
Besides, we record the gradient compression ratios of these devices with the \textsf{CAC} approach.
By Figure \ref{fig:motivation_d}, it can be found that the important devices often adopt large gradient compression ratios, which introduce significant errors on their critical local gradients, hindering the global model from learning the important knowledge.

In summary, under the challenges of data heterogeneity and model obsolescence, the existing approaches often cause non-negligible model/gradient compression errors.
This limitation motivates us to design a low-deviation compression approach that can reduce traffic costs efficiently while minimizing the impact on model accuracy.

\vspace{-1mm}
\section{System Overview of Caesar}\label{sec:overview}
Motivated by the limitation of existing approaches, we propose a novel FL system, named \method, which \uline{\textbf{c}}ompresses the gr\uline{\textbf{a}}diens and mod\uline{\textbf{e}}ls in tran\uline{\textbf{s}}mission via a low-devi\uline{\textbf{a}}tion app\uline{\textbf{r}}oach.
In this section, we will provide an overview of \method's workflow.

\begin{figure}[t]\centering
    \includegraphics[width=1.0\linewidth]{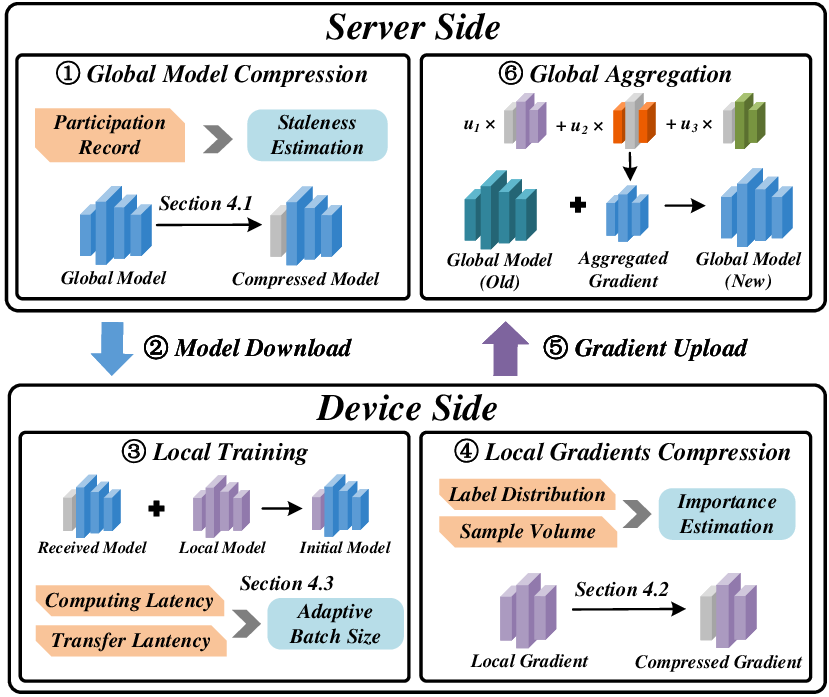}
    \caption{Overview of \method's workflow.}\label{fig:framework}
    \vspace{-5mm}
 \end{figure}

Different from the typical FL paradigm \cite{mcmahan2017communication}, there are six main steps in \method's workflow, illustrated in Figure \ref{fig:framework}.
Specifically, at the beginning of each round, a portion of the devices are selected to participate in training.
Next, the PS adopts the devices' participation record to quantify their local models' staleness, which guides the determination of the global model compression ratio (\circled{1}).
After that, the PS dispatches the compressed global models to corresponding participants (\circled{2}).
Once receiving the model, each device recovers the original global model via its local model.
Since the model/gradient compression ratios vary across devices, their transmission latencies differ accordingly.
This disparity can become more pronounced when devices have heterogeneous bandwidth.
Moreover, the computing power of devices also varies, leading to different training latencies (\ie, time cost for one training iteration).
As a result, the faster devices must wait for the slower ones to complete global aggregation, reducing training efficiency.
To tackle this problem, each participant performs local training with an adaptive batch size (\circled{3}).
For example, the devices with low computing and/or transmission latency are configured with large batch sizes, and vice verse, ensuring that the round time is essentially the same across participants.
After local training, each participant will derive the local gradient, whose importance is estimated by the local data properties, including label distribution and sample volume.
Based on the estimated importance, the devices compute the local gradient compression ratios and compress their local gradients (\circled{4}).
Subsequently, each device uploads its compressed local gradient to the PS (\circled{5}).
Once receiving the local gradients from all participants, the PS performs global aggregation.
Finally, the global model is updated according to the aggregated gradient (\circled{6}).
Such a communication round will be conducted repeatedly until the global model achieves the target accuracy.
Notably, we impose no limitation on device selection strategy or weight assignment algorithm, making it compatible with a mass of recent FL system literatures.

\section{System Design of Caesar}\label{sec:design}
\subsection{Global Model Compression}
This component aims to improve the fidelity of each device's initial model for local training by optimizing the process of global model compression.
Firstly, the PS determines a suitable global model compression ratio for every participant.
As discussed above, the precision of the restored model depends on both the local model staleness and the global model compression ratio.
Specifically, for devices that participate in training infrequently, their local models tend to become obsolete. 
Consequently, these devices may struggle to accurately recover the received model when a large compression ratio is applied. 
Conversely, devices with high participation frequencies typically maintain fresher local models, allowing them to recover the received model effectively even with a large model compression ratio.
Inspired by this insight, we estimate the staleness of each device's local model by its participation history, \ie, the number of missed communication rounds.
For example, if the last participation of device $v_{i}$ occurred in round $r_{i}$ $\in \{0, 1, \cdots, t - 1 \}$, its model staleness at round $t$ ($t > r_{i}$) is $\delta_{i}^{t} = t - r_{i}$, where $r_{i} = 0$ indicates that the device has never been selected for training.
Intuitively, a higher $\delta_{i}^{t}$ indicates a more outdated local model and, correspondingly, a weaker model recovery capability, and vice versa.
Based on this regular pattern, we design a staleness-aware method for determining the global model compression ratio, which is formulated as follows:
\begin{equation}\label{eq:d_ratio}
    \theta_{d, i}^{t} = (1- \frac{\delta_{i}^{t}}{t}) \cdot \theta_{d}^{max}, \quad \forall v_{i} \in \mathbf{N}^{t}, \forall t \in [T]
\end{equation}
where the positive $\theta_{d}^{max}$ is the upper bound of the global model compression ratio.
In this way, the higher the device’s staleness, the lower the compression ratio applied during the model download, ensuring that the device receives a more accurate global model for recovery. Consequently, each device is equipped with a well-initialized model for local training.
Note that if a device $v_{i}$ has never participated in training (\ie, $r_{i} = 0$), its local model is unavailable.
In this case, its staleness $\delta_{i}^{t}$ equals $t$ and it will receive the full-precise model with a compression ratio of $\theta_{d, i}^{t} = 0$.

\begin{figure}[t]\centering
    \includegraphics[width=1.0\linewidth]{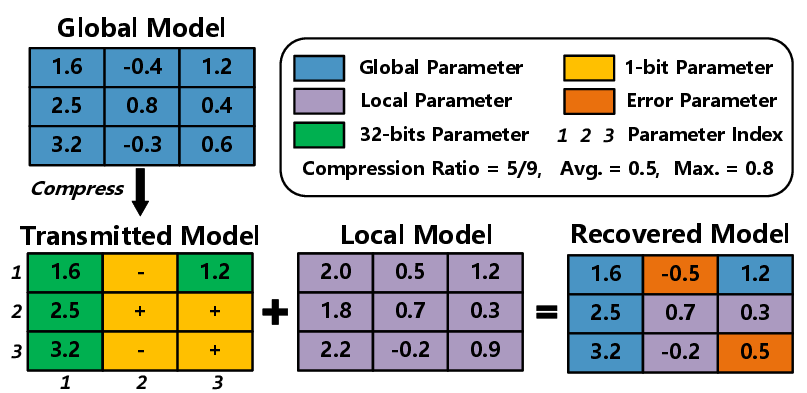}
     \vspace{-3mm}
    \caption{Illustration of the model recovery mechanism in \method with a compression ratio of 5/9.} \label{fig:model_recovery}
        \vspace{-3mm}
 \end{figure}

Secondly, the global model will be compressed for each device $v_{i}$ with the compression ratio $\theta_{d, i}^{t}$.
To mitigate the information loss and enhance the device’s ability to recover the received model, we propose a novel model compression method on the basis of Top-K sparsification \cite{alistarh2018convergence} and 1-bit quantization \cite{bernstein2018signsgd}.
Specifically, for each device $v_{i}$, the PS first selects the $\theta_{d, i}^{t}$ proportion of elements (\ie, parameters) with the smallest absolute values, and represents them by 1 bit.
In other words, only the signs of these elements are sent to the devices.
The remaining $(1 - \theta_{d, i}^{t})$ proportion of elements are sent with full precision (typically 32 bits).
Besides, the maximum and average absolute values of the elements quantized to 1 bit will also be transmitted to the devices, facilitating subsequent model recovery.
As illustrated in Figure \ref{fig:model_recovery}, with a compression ratio of 5/9, 4 elements with the largest absolute values (\ie, green rectangles) maintain the 32 bits precision, while the remaining 5 elements (\ie, yellow rectangles) are represented as 1 bit for transmission, and their average and maximum absolute values are 0.5 and 0.8, respectively.

Thirdly, devices will execute the model recovery once receiving the models from the PS.
Specifically, each device tries to approximate the 1-bit elements in the received model using the corresponding 32-bit elements in its previous local model.
As shown in Figure \ref{fig:model_recovery}, the elements in position (2, 2), (2, 3), and (3, 2) are replaced by those in the same position of the local model, where position ($x$, $y$) represents row $x$ and column $y$ in the matrix.
Since the local model has some similarity with the global model, such parameter replacement can effectively reduce the model deviation caused by compression.
However, there are still two kinds of obvious errors for the local parameters in this approximation process.
(1) The sign bit is opposite to the original one.
(2) The absolute value exceeds the expected maximum.
By Figure \ref{fig:model_recovery}, the local parameters in positions (1, 2) and (3, 3) separately show the two types of errors.
To address this, we approximate these parameters by the average absolute value with the original sign, instead of the local parameters.
In the restored model shown in Figure \ref{fig:model_recovery}, the parameters in positions (1, 2) and (3, 3) are approximated as -0.5 and 0.5, respectively.

Nevertheless, there is a potential concern in this component.
The PS needs to compress the global model for each participant using a customized ratio, which incurs substantial computational overhead as the number of participants increases.
To address this issue, we propose a cluster-based solution.
At the beginning of each round $t$, participants are grouped into $K$ clusters based on their model staleness, where devices in the same cluster exhibit similar levels of local model staleness. 
For each cluster, the PS calculates an average staleness value and determines a single compression ratio to be applied to all devices within that cluster. 
In this way, the PS only needs to conduct the model compression for $K$ times, instead of $|\mathbf{N}^{t}|$ times, in each round $t$.
Obviously, increasing the number of clusters improves the accuracy of the compression ratios assigned to devices but also raises the computational burden on the PS. 
The number of clusters $K$ can be adjusted flexibly to achieve a balance between computational efficiency and model recovery precision based on specific system requirements.

\subsection{Local Gradient Compression} \label{sec:gradient_compression_component}
This component adjusts the local gradient compression ratios for different devices, so as to diminish the deviation on important gradients.
Due to data heterogeneity, both the label distribution and the sample volume are various across devices.
Therefore, the local gradients from diverse devices may have different importance for the global model convergence.
Specifically, the devices with data distributions closely aligned with the global distribution are typically more critical because their local gradients point in the correct optimization direction \cite{liao2024mergesfl}.
Besides, the devices with rich sample volume can be valuable as they contribute more knowledge to the global model.
On the one hand, the less critical local gradients can be compressed with higher ratios to reduce network traffic. 
On the other hand, lower gradient compression ratios should be applied to more important devices to better preserve their contributions to global model convergence.

Based on the above analysis, we propose an adaptive gradient compression approach.
Concretely, before the FL training, the PS first estimates the importance $\mathbb{C}_{i}$ of each device $v_{i} \in \mathbf{N}$, which depends on two aspects. 
(1) The sample volume $A_{i}$.
Specifically, a larger sample volume $A_{i}$ means that device $v_{i}$ can provide more data samples for model training, leading to richer knowledge in its local gradient $g_{i}^{t}$ and, consequently, higher importance $\mathbb{C}_{i}$.
(2) The gap between local data distribution $\Phi_{i}$ and global data distribution $\Phi_{0}$, denoted as $D_{i}$.
Considering a $H$-classification task, device $v_{i}$'s local data distribution $\Phi_{i}$ can be represented as a vector $[e_{i}^{1}, e_{i}^{2}, \cdots, e_{i}^{H}]$, where $e_{i}^{h} \in [0, 1]$ ($h \in [H]$) is the proportion of data with the $h$-th label in the whole dataset $\mathcal{D}_{i}$, and $\sum_{h=1}^{H} e_{i}^{h} = 1$.
The global data distribution is typically uniform with $e_{0}^{h} = \frac{1}{H}$ ($\forall h$).
We then adopt the KL-divergence \cite{hershey2007approximating, goldberger2003efficient} to measure the distribution gap $D_{i}$ as follows:
\begin{equation}
    D_{i} = KL(\Phi_{i} || \Phi_{0}) = \sum_{h = 1}^{H} e_{i}^{h} \cdot \ln \frac{e_{i}^{h}}{e_{0}^{h}}, \quad \forall v_{i} \in \mathbf{N}.
\end{equation}

For a device $v_{i}$, the lower $D_{i}$ indicates that the direction of the local gradient $g_{i}^{t}$ is closer to the optimal direction, which in turn increases the device’s importance $\mathbb{C}_{i}$.
Here, we can estimate each device $v_{i}$'s importance based on $A_{i}$ and $D_{i}$:
\begin{equation} \label{eq:con}
    \mathbb{C}_{i} = \lambda \cdot \frac{A_{i}}{A^{max}} + (1 - \lambda) \cdot \frac{1}{\text{e}^{D_{i}}}, \quad \forall v_{i} \in \mathbf{N}^{t}
\end{equation}
where $A^{max}$ is the maximum threshold for the devices' local sample volume.
The weights of $A_{i}^{t}$ and $D_{i}$ is regulated by the hyper-parameter $\lambda \in [0, 1]$, which is set to 0.5 by default.

Inspired by \pyramidfl \cite{li2022pyramidfl}, we utilize a rank-based strategy to determine proper local gradient compression ratios for different participants.
Specifically, the PS will rank all devices by their importance in descending order and compute the gradient compression ratio of device $v_{i}$ based on its ranking in each round $t$ as follows:
\begin{equation} \label{eq:u_ratio}
    \theta_{u, i}^{t} = \theta^{min}_{u} + \frac{\theta^{max}_{u} - \theta^{min}_{u}}{| \mathbf{N} |} \cdot Rank(\mathbb{C}_{i})
\end{equation}
where $\theta^{min}_{u}$ and $\theta^{max}_{u}$ are separately the low and up bounds of gradient compression ratio.
In summary, the important devices will adopt a small gradient compression ratio, so as to preserve the value of their important local gradients.
Notably, this ranking-based strategy requires the PS to collect the importance values from all devices before the FL training.
However, there are no privacy concerns associated with sharing these important values, as neither the exact data volumes nor label distributions can be inferred from them \cite{ye2023feddisco}.
Moreover, this adaptive approach is method-agnostic, allowing the integration of various popular compression techniques, such as sparsification, quantization, and tensor decomposition, among others.

\subsection{Batch Size Optimization}

Considering the variations in devices' participation frequency and local data properties, \method adjusts model/gradient compression ratios for different participants.
Consequently, communication times can vary significantly across devices, leading to non-negligible idle waiting periods for faster devices. 
This issue greatly reduces the training efficiency in FL, resulting in excessive training time.
As studied in many previous works \cite{liao2024mergesfl, ma2023adaptive}, a larger batch size enables the model to learn more data at each iteration, thereby accelerating the convergence, but it also results in longer computing latency, and vice versa.
Inspired by this observation, we design a greedy-based strategy to optimize devices' batch size for model training.
Specifically, the fast devices utilize large batch sizes for better training performance, while the slow devices adopt the small batch sizes to advance the global aggregation, so as to accelerate the FL training.

For clarity of explanation, we first formulate the time cost of device $v_{i}$ in each round $t$ as follows:
\begin{align} \label{eq:duration_time}
    M_{i}^{t} &= M_{d, i}^{t} + M_{u, i}^{t} + M_{c, i}^{t}  \notag \\
    &= \theta_{d, i}^{t} \cdot (Q / \beta_{d, i}^{t}) + \theta_{u, i}^{t} \cdot (Q / \beta_{u, i}^{t}) + \tau \cdot b_{i}^{t} \cdot \mu_{i}^{t}
\end{align} \label{eq:round_time}
For the participant $v_{i}$ in round $t$, the first term $M_{d, i}^{t}$ and the second term $M_{u,i}^{t}$ separately denote the model download time and the gradient upload time, where $Q$ is the size of the uncompressed model/gradient.
$\beta_{d, i}^{t}$ and $\beta_{u, i}^{t}$ represent the download and upload bandwidth, respectively.
The third term $M_{c,i}^{t}$ is the computation time for model training, where $\tau$ is the number of local iterations, 
$b_{i}^{t}$ is the batch size, and $\mu_{i}^{t}$ is the time cost for processing one data sample.
In synchronous FL, the completion time $M^{t}$ of each round $t$ depends on the slowest participant, \ie $M^{t} = \max_{v_{i} \in \mathbf{N}^{t}} M_{i}^{t}$. 

To optimize the batch size, the PS first selects the fastest device in each round $t$ (denoted as $v_{l}^{t}$) as follows:
\begin{equation}  \label{eq:fast}
    v_{l}^{t} = \min_{v_{i} \in \mathbf{N}^{t}} (M_{d, i}^{t} + M_{u, i}^{t} + \tau \cdot b^{max} \cdot \mu_{i}^{t})
\end{equation} 

$b^{max}$ represents a predefined maximum batch size.
For this device $v_{l}^{t}$, the PS assigns $b^{max}$ to it.
For each remaining participant $v_{i} \in \mathbf{N}^{t} - \{v_{l}^{t}\}$, we adjust its batch size $b_{i}^{t}$ to ensure its round time cost $M_{i}^{t}$ dose not exceeds that of the fastest device $M_{l}^{t}$:
\begin{align} \label{eq:batch_size}
    b_{i}^{t} = \lfloor \frac{M_{l}^{t} - M_{d, i}^{t} - M_{u, i}^{t}}{\tau \cdot \mu_{i}^{t}} \rfloor
\end{align}

With the above fine-grained batch size configuration, the completion time for each round $t$ depends on the fastest participant $v_{l}$, thereby the training process can be accelerated.
Besides, the appropriate batch size configurations make the duration time of other devices close to that of the fastest device, which can effectively diminish the idle waiting time.

\begin{algorithm}[t] 
\caption{Training Process of \method.} \label{alg}
\KwIn{Device set $\mathbf{N}$, participation rate $\alpha$, compression ratios' bounds $\theta_{d}^{max}$, $\theta_{u}^{min}$ and $\theta_{u}^{max}$, maximum batch size $b^{max}$, the number of local iteration $\tau$, target model accuracy $Acc_{target}$, learning rate $\eta^{0}$.}
\KwOut {A model that achieves the target accuracy}
\textbf{Server Executes:} \\
Set $t \leftarrow 0$, $Acc \leftarrow 0$ \\ \label{alg:init1}
Initialize the global model $w^{0}$ \\ 
Rank devices by the importance $\mathbb{C}_{i}$, $\forall v_{i} \in \mathbf{N}$ \\ \label{alg:init2}
\While {$Acc < Acc_{target}$}{ \label{alg:round1}
    Select the participant set $\mathbf{N}^{t}$ from $\mathbf{N}$ with rate $\alpha$ \\
    \For {each device $v_{i} \in \mathbf{N}^{t}$}{
        Compute the ratio $\theta_{d, i}^{t}$ by Eq. \eqref{eq:d_ratio} \\ \label{alg:config1}
        Compute the ratio $\theta_{u, i}^{t}$ by Eq. \eqref{eq:u_ratio} \\
        Compute the batch size $b_{i}^{t}$ by Eq. \eqref{eq:batch_size} \\ \label{alg:config2}
        Compress global model $\overline{w}^{t}_{i} \leftarrow S(w^{t}, \theta_{d, i}^{t})$ \\
        $\overline{g}_{i}^{t}$ $\leftarrow$ DeviceUpdate($\overline{w}^{t}_{i}$, $b_{i}^{t}$, $\theta_{u, i}^{t}$, $\tau$, $\eta^{t}$) \\
    }
    Update global model $w^{t + 1} \leftarrow w^{t} - \sum_{v_{i} \in \mathbf{N}^{t}} \frac{1}{\| \mathbf{N}^{t} \|} \overline{g}_{i}^{t}$ \\ \label{alg:agg2}
    $Acc \leftarrow$ (evaluate $w^{t + 1}$) \\ \label{alg:test}
    Set $t \leftarrow t + 1$ \\ 
     \label{alg:round2}
}
\textbf{Return} global model $w^{t}$ \\
\textbf{DeviceUpdate($\overline{w}^{t}_{i}$, $b_{i}^{t}$, $\theta_{u, i}^{t}$, $\tau$, $\eta^{t}$):} \\
Recover the received model via the local model $w_{i}$  \\ \label{alg:re1}
Initialize the model $w_{i}^{t, 0} \leftarrow \widehat{w}_{i}^{t}$ (the restored model) \\ \label{alg:re2}
\For {each iteration $j = \{0, 1, \cdots, \tau - 1 \} $}{  \label{alg:train1}
    Get a random data batch $\zeta_{i}^{t, j}$ with size $b_{i}^{t}$ from $\mathcal{D}_{i}$ \\
    $w_{i}^{t, j + 1} \leftarrow w_{i}^{t, j} - \eta^{t} \nabla l(w_{i}^{t,j}; \zeta_{i}^{t, j})$ \\ \label{alg:train2}
}
Derive the local gradient $g_{i}^{t} = \eta^{t} \sum_{j = 0}^{\tau - 1} 
 \nabla l(w_{i}^{t, j}; \zeta_{i}^{t, j})$ \\
Update the local model $w_{i} \leftarrow w_{i}^{t, \tau}$ \\
Compress the local gradients $\overline{g}_{i}^{t} \leftarrow S(g_{i}^{t}, \theta_{u, i}^{t})$ \\ \label{alg:upload2}
\textbf{Return} $\overline{g}_{i}^{t}$ to server

\end{algorithm}

\subsection{Training Process of \method}
In this section, we integrate the two components into a complete FL framework, \method.
The training process of \method is summarized in Algorithm \ref{alg}.
To begin with, the developer inputs the system parameters and sets a target model accuracy.
Next, the PS performs initialization (Lines \ref{alg:init1}-\ref{alg:init2}) and starts the first communication round.
In each round, the PS first determines the optimal model and gradient compression ratios, as well as the batch size for each participant (Lines \ref{alg:config1}-\ref{alg:config2}).
Let $S(\cdot)$ denote the compression executor.
Then, the PS compresses the global model and dispatches it to the corresponding participant, along with the training parameters (\ie, batch size, the number of local iterations, learning rate, and gradient compression ratio).
After that, each selected device attempts to recover the compressed global model by its previous local model, and adopts the recovered model as the initial model for local training (Lines \ref{alg:re1}-\ref{alg:re2}).
Next, the devices conduct the local iteration for $\tau$ times with the adaptive batch size configured by the PS (Lines \ref{alg:train1}-\ref{alg:train2}).
Once completing the training, each participant compresses its local gradient and uploads the compressed gradients to the PS.
Once collecting the gradients from all participants, the PS aggregates the local gradients to update the global model (Line \ref{alg:agg2}).
Finally, the updated global model will be evaluated on the test dataset to obtain the current test accuracy (Line \ref{alg:test}).
Such a communication round (Lines \ref{alg:round1}-\ref{alg:round2}) is conducted iteratively until the global model achieves the predefined target accuracy.

\section{System Implementation} \label{sec:implementation}

\begin{figure}[t]\centering
    \includegraphics[width=1.0\linewidth]{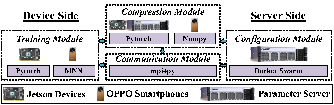}
    \caption{The system architecture of \method.} \label{fig:implememtation}
    \vspace{-3mm}
 \end{figure}

In order to evaluate the performance of \method, we utilize Python to implement an FL system on two physical prototypes (\ie, the NVIDIA Jetson and OPPO smartphones systems), which will be introduced in detail in Section \ref{sec:exp_set}.
Specifically, our implementation is composed of the following four modules, including about 3k lines of Python code.

\textbf{Training Module} supports the model training on devices. 
For the Jetson kits, we adopt PyTorch \cite{paszke2019pytorch} to implement the training module with the environment of Ubuntu 18.04, CUDA v10.0, and cuDNN v7.5.0.
The training module on OPPO smartphones is supported by the MNN engine \cite{jiang2020mnn}, where the operation system is the ColorOS 12\footnote{https://www.oppo.com/en/coloros12/}.

\textbf{Configuration Module} is responsible for the training parameters configuration (\ie, model/gradient compression ratios, batch size).
This module is based on the docker technique, and each device is abstracted as a docker container.
We utilize the Docker Swarm \cite{naik2016building} to detect each device's status and information (\eg, bandwidth, training latency), allowing the module to make suitable configurations timely.

\textbf{Communication Module} establishes the connection between devices and the PS.
We implement the communication module by the mpi4py package \cite{dalcin2021mpi4py}, which provides a collection of sending and receiving interfaces (\eg, \textit{mpi4py.MPI.Comm.send}, \textit{mpi4py.MPI.Comm.recv}), facilitating more efficient parallel communication among containers.

\textbf{Compression Module} executes the model and gradient compression on devices and the PS.
In \method, we mainly focus on the compression technique of Top-K sparsfication.
For the Jetson devices and the PS, we implement the compression module using the \textit{torch.topk} API, which can enjoy the GPU acceleration.
For the OPPO smartphones without GPU, the sparsification function is implemented based on the Numpy package \cite{harris2020array}.

\section{Performance Evaluation}\label{sec:evaluation}
\subsection{Experimental Methodology} \label{sec:exp_set}

\textbf{Applications, Datasets, and Models.} To demonstrate \method's generality across various datasets and models, we evaluate the performance of \method on the following four representative AI applications, which are widely deployed on mobile devices \cite{wang2023bose}.

\textbf{1) Image Classification} is a popular computer vision application on mobile devices.
In this work, we train the ResNet-18 model \cite{he2016deep} over the CIFAR-10 dataset \cite{krizhevsky2009learning} for evaluation.
Specifically,  CIFAR-10 includes 60,000 32$\times$32 color images evenly from 10 categories, with 50,000 for training and 10,000 for testing.

\textbf{2) Human Activity Recognition} is to identify a person's status (\eg, standing, sitting) based on the sensor data from smartphones or smartwatches. 
For this task, we adopt the HAR dataset \cite{anguita2013public} collected from 30 individuals, including 7,352 training samples and 2,947 testing samples across 6 categories.
The model trained on HAR is a CNN model (denoted as CNN-H) with three 5$\times$5 convolutional layers and two fully-connected layers \cite{liao2024mergesfl}.

\textbf{3) Speech Recognition} is an application that recognizes voice commands from target words (\eg, "yes", "left", "seven"). 
We adopt the Google Speech dataset \cite{warden2018speech} for this application.
The dataset includes 85,511 and 4,890 audio clips for training and testing, respectively, which are categorized into 35 classes by the corresponding keyword.
A CNN model (denoted as CNN-S) with four 1-D convolutional layers and one fully-connected layer \cite{wang2023bose} is trained on this dataset.

\textbf{4) E-Commerce Recommendation} predicts whether a customer will be interested in a product according to his/her historical click behaviors.
For this application, we adopt an industrial dataset built from the click logs of 8,102 OPPO theme store\footnote{https://play.google.com/store/apps/details?id=com.heytap.themestore} users, denoted as OPPO-TS.
OPPO-TS is collected from June, 2023 to November, 2023, containing about 90 thousand training samples and 10 thousand testing samples in total.
A logistic regression (LR) model with 129,314 features is trained and tested over this dataset.

\noindent \textbf{System Implementation.}
In order to conduct experiments for performance evaluation, we have implemented \method in two physical prototype systems.

\textbf{1) NVIDIA Jetson System} is composed of 80 NVIDIA Jetson devices\footnote{https://docs.nvidia.com/jetson}, including 30 TX2 devices, 40 NX devices, and 10 AGX devices.
Each TX2 is equipped with a 256-core Pascal GPU and a CPU cluster consisting of a 2-core Denver2 and a 4-core ARM.
Each NX (or AGX) is outfitted with a 384-core (or 512-core) Volta GPU and a 6-core (or 8-core) Carmel CPU CortexA57. 
The detailed technical specifications of these Jetson devices are listed in Table \ref{table:jetson}.
Experiments of image classification, human activity recognition, and speech recognition are conducted on this prototype system.

\textbf{2) OPPO Smartphone System} is constructed by 40 OPPO smartphones.
Specifically, there are 15 A1, 15 Reno9, and 10 FindX6.
The SoCs equipped on A1, Reno9, and FindX6 are Qualcomm Snapdragon 695, 778G, and 8Gen2, respectively.
Table \ref{table:oppo} summarizes the detailed performance indicators of these three types of OPPO smartphones.
Experiments of the e-commerce recommendation application are conducted on this prototype system.

In these two systems, we employ an AMAX deep learning workstation as the PS, which is equipped with an Intel(R) Xeon(R) Platinum 8358P CPU, four NVIDIA RTX A6000 GPUs, and 512GB RAM.

\begin{table}[t]
\caption{Technical specifications of NVIDIA Jetson Kits.}
\label{table:jetson}
\centering
\begin{tabular}{c|ccc}
\hline
\textbf{Types} & \tabincell{c}{\textbf{AI} \\ \textbf{Performance}} & \textbf{RAM} & \tabincell{c}{\textbf{CPU/GPU} \\ \textbf{Frequency}} \\ 
\hline
TX2 & 1.33 TFLOPs & 8GB LPDDR4 & 2.0/1.12GHz \\ 
NX  & 21 TOPs & 8GB LPDDR4x & 1.9/1.1GHz \\ 
AGX  & 32 TOPs & 32GB LPDDR4x & 2.2/1.37GHz\\ 
\hline
\end{tabular}
\end{table}

\begin{table}[t]
\caption{Technical specifications of OPPO smartphones.}
\label{table:oppo}
\centering
\begin{tabular}{c|ccc}
\hline
\textbf{Types} & \tabincell{c}{\textbf{AI} \\ \textbf{Performance}} & \textbf{RAM} & \ \tabincell{c}{\textbf{SoC} \\ \textbf{Frequency}} \\ 
\hline
A1 & 486.4 GFLOPs & 8GB LPDDR4x & 2.2GHz \\ 
Reno8 & 844 GFLOPs & 12GB LPDDR5 & 2.4GHz \\ 
FindX6  & 3481.6 GFLOPs & 16GB LPDDR5x & 3.36GHz\\ 
\hline
\end{tabular}
\end{table}

\begin{table*}[t]
  \centering
  \caption{Summary of \method's performance improvement over the four baselines.} \label{table:summary}
  \begin{tabular}{cc|cc|cc|cc|cc|cc}
    \hline
    \multirow{2}{*}{Datasets} & Target & \multicolumn{2}{c|}{\fedavg} & \multicolumn{2}{c|}{\flexcom} & \multicolumn{2}{c|}{\prowd} & \multicolumn{2}{c|}{\pyramidfl} &  \multicolumn{2}{c}{\method} \\
    & Acc./AUC &  Traffic & Time & Traffic & Time & Traffic & Time & Traffic & Time & Traffic & Time  \\\hline
    CIFAR-10 & 0.80 & 219.15GB & 5.17h & 192.35GB & 4.41h & 160.24GB & 3.87h & 187.56GB & 3.48h & \textbf{115.57GB} & \textbf{2.76h} \\\hline
    HAR & 0.86 & 44.99GB & 6.21h & 37.09GB & 4.83h & 31.89GB & 3.86h & 35.34GB & 3.47h & \textbf{24.26GB} & \textbf{2.37h} \\\hline
    Speech & 0.87 & 520.84MB & 4.02h & 429.36MB & 3.38h & 357.87MB & 2.99h & 396.69MB & 2.46h & \textbf{269.98MB} & \textbf{1.95h}  \\\hline
    OPPO-TS & 0.65 & 5.24GB & 1.03h & 3.89GB & 1.98h & 2.81GB & 2.24h & 2.96GB & 1.12h &  \textbf{1.83GB} & \textbf{0.86h} \\\hline
  \end{tabular}
\end{table*}

\noindent \textbf{Setting of System Heterogeneity.}
In real-world situations, the computation and communication capabilities of mobile devices are typically heterogeneous and dynamic.
To simulate this property, we configure the devices as follows.

\textbf{1) For Computation.}
The devices can be configured to work with different modes, specifying processing speed and energy consumption.
The configuration of TX2 offers four modes, while that of NX and AGX provides eight modes.
OPPO smartphones support two work modes, \ie, normal and power-saving.
Therefore, the training speed varies significantly across devices, reaching up to 100 times difference (\eg, mode 0 of AGX and mode 1 of TX2).
To further reflect the time-varying on-device resources, we randomly change the modes of devices every 20 communication rounds \cite{wang2023bose}.

\textbf{2) For Communication.}
All devices are connected to the PS via a WiFi router.
We follow \cite{liao2024mergesfl} to divide the devices into four groups evenly and place these groups in different rooms, which are about 2m, 8m, 14m, and 20m away from the WiFi router.
Due to random channel noise and device competition, the bandwidth between devices and the PS will vary dynamically.
Specifically, the measured bandwidth fluctuates within the range of [1Mb/s, 30Mb/s] approximately.

\noindent\textbf{Setting of Data Heterogeneity.}
In the experiments about CIFAR-10, HAR, and Speech, we create the heterogeneous local datasets based on Dirichlet distributions \cite{hsu2019measuring, yurochkin2019bayesian}.
Specifically, the training samples of each client are drawn independently by a vector $\mathbf{v}$ $\sim Dir(\delta \mathbf{q})$, where $\mathbf{q}$ characterizes a prior class distribution, and $\delta > 0$ indicates the identicalness among devices \cite{liao2024mergesfl}.
We define $p = \frac{1}{\delta}$ to quantify the data heterogeneity, where a higher $p$ means a higher statistic heterogeneity degree.
Given $p > 0$, both data volume and data distribution will be various across different devices. 
$p = 0$ is a special case, representing IID with an identical volume for all devices' datasets.
For the OPPO-TS dataset, we naturally assign the training samples to devices by the user's ID, \ie, a device represents a real user.
Therefore, the data heterogeneity level for OPPO-TS is fixed.
Our experiments are conducted with five different data heterogeneity levels (\ie, $p$=1, 2, 4, 5, and 10), and the default setting of $p$ is 5.

\begin{figure*}[t]
\centering
\subfigure[CIFAR-10]
{
    \includegraphics[width=0.22\linewidth,height=3.2cm]{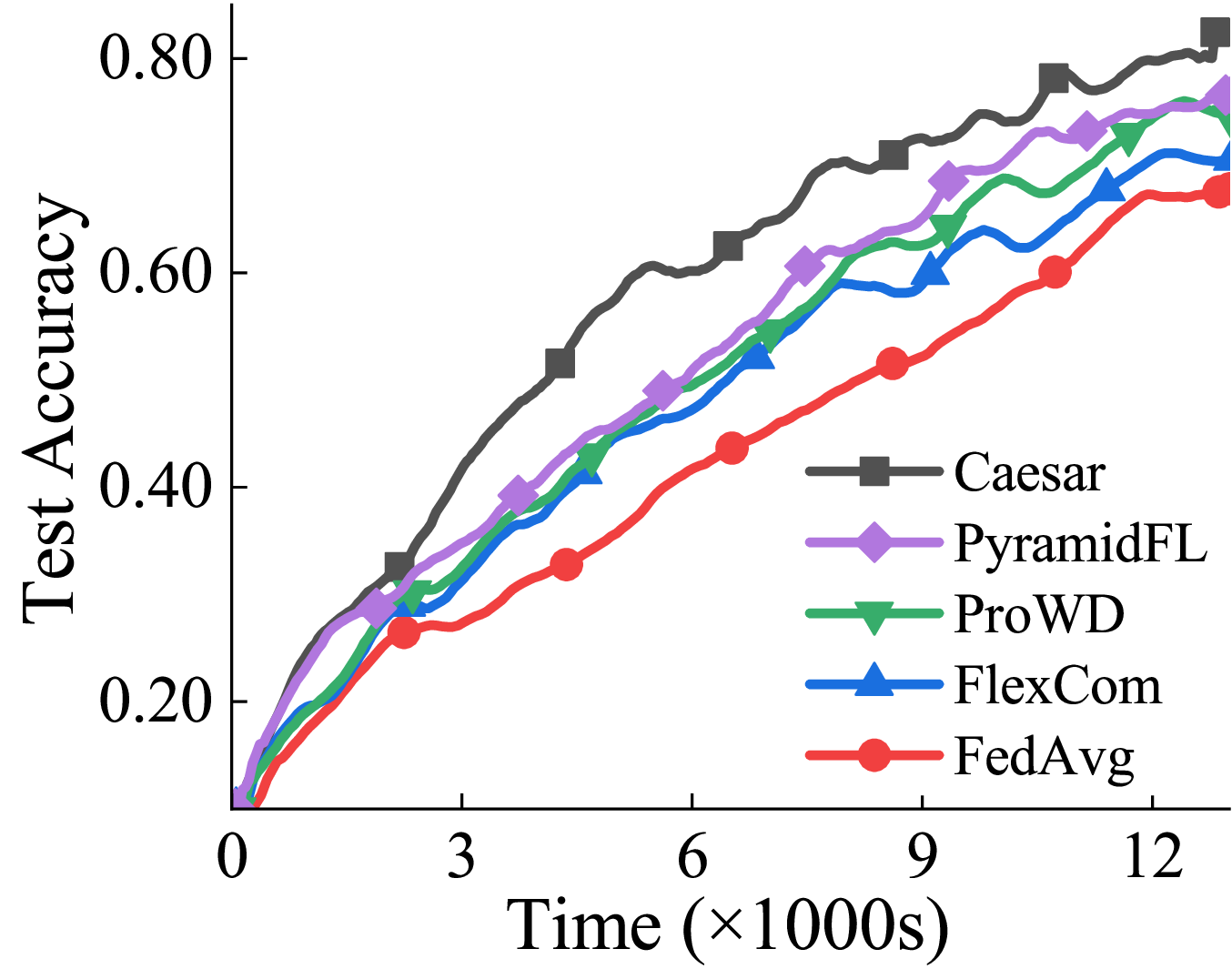}
    \label{fig:cifar_time_acc}
}\quad 
\subfigure[HAR]
{
    \includegraphics[width=0.22\linewidth,height=3.2cm]{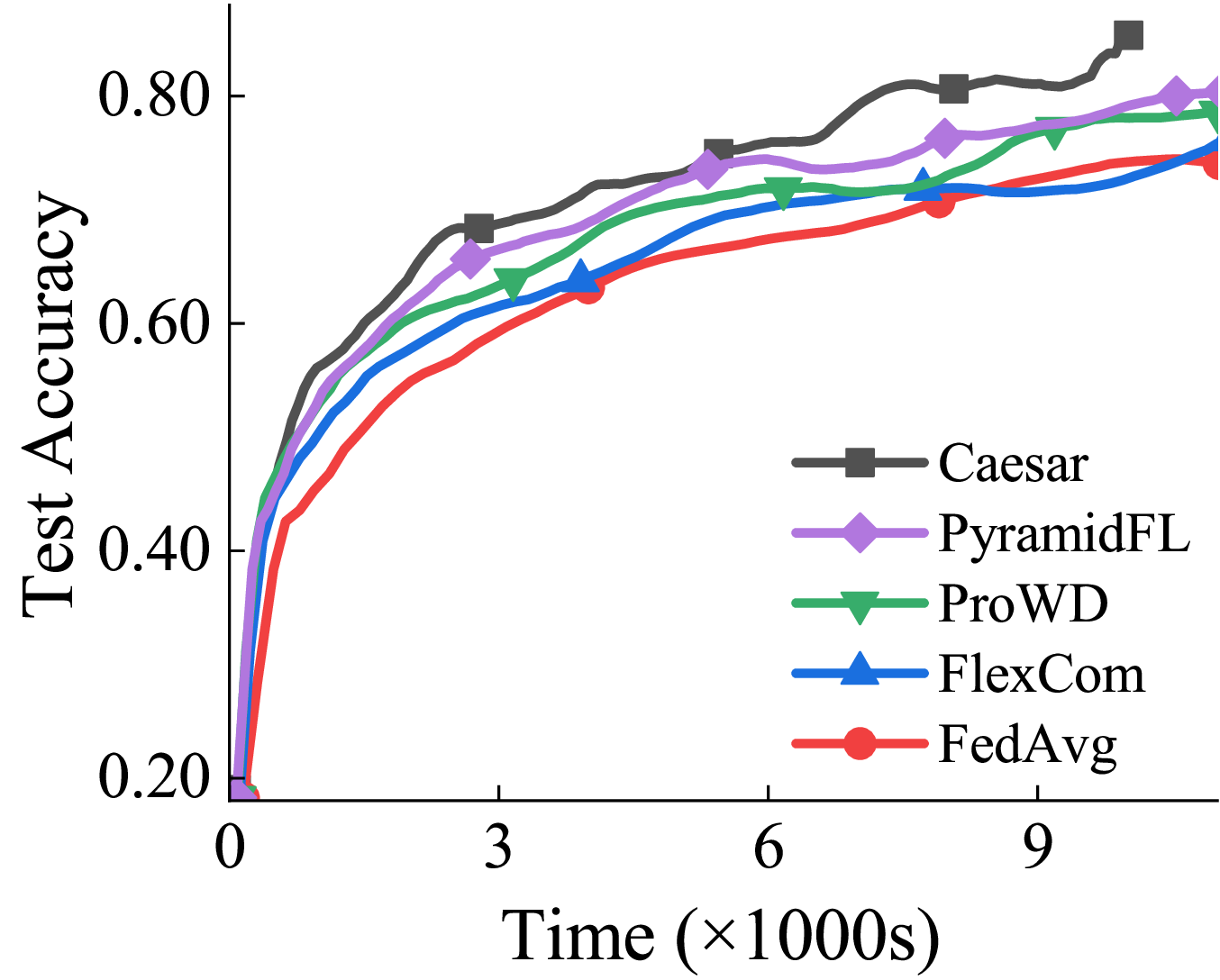}
    \label{fig:har_time_acc}
}\quad 
\subfigure[Speech]
{
    \includegraphics[width=0.22\linewidth,height=3.2cm]{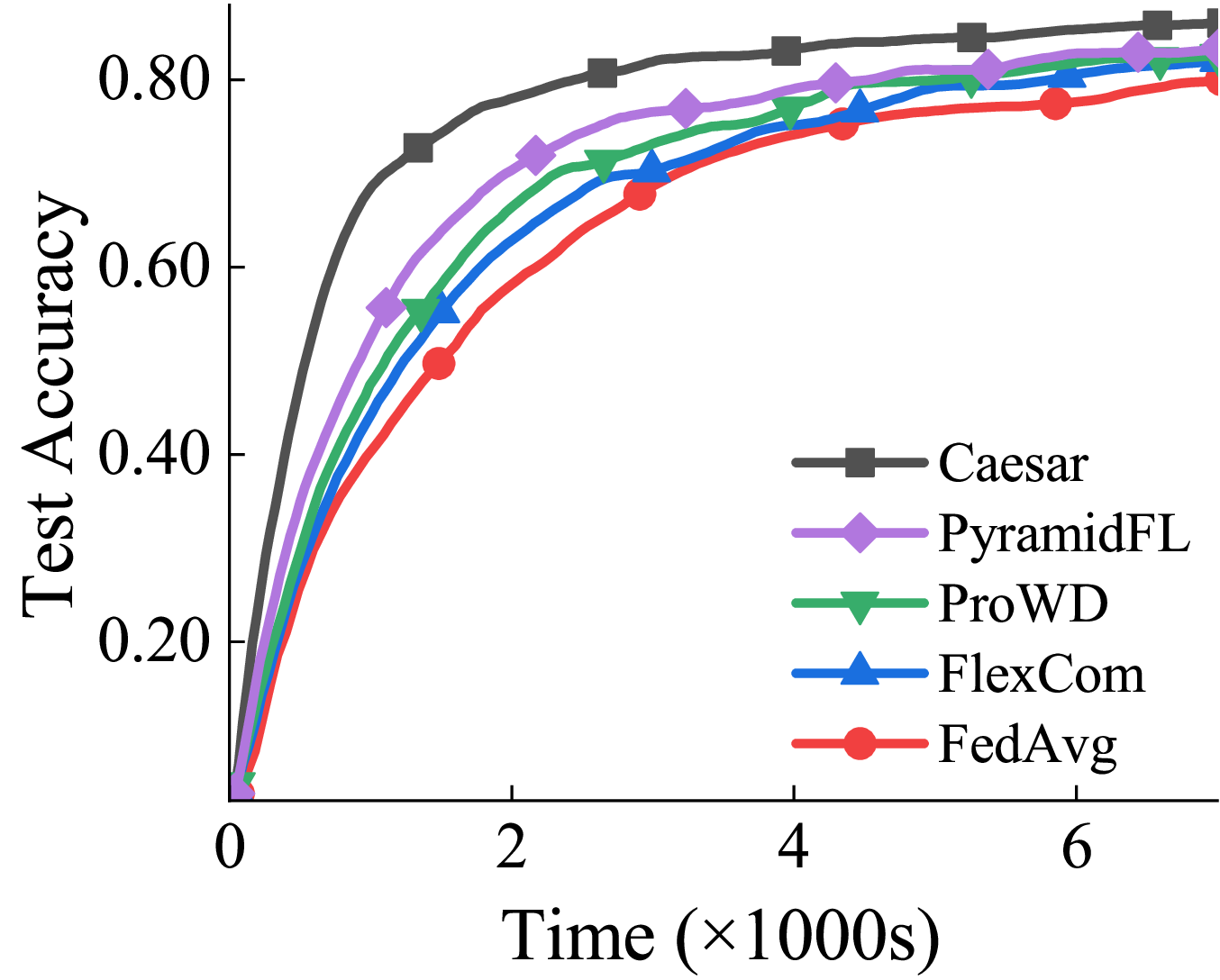}
    \label{fig:speech_time_acc}
}\quad 
\subfigure[OPPO-TS]
{
    \includegraphics[width=0.22\linewidth,height=3.2cm]{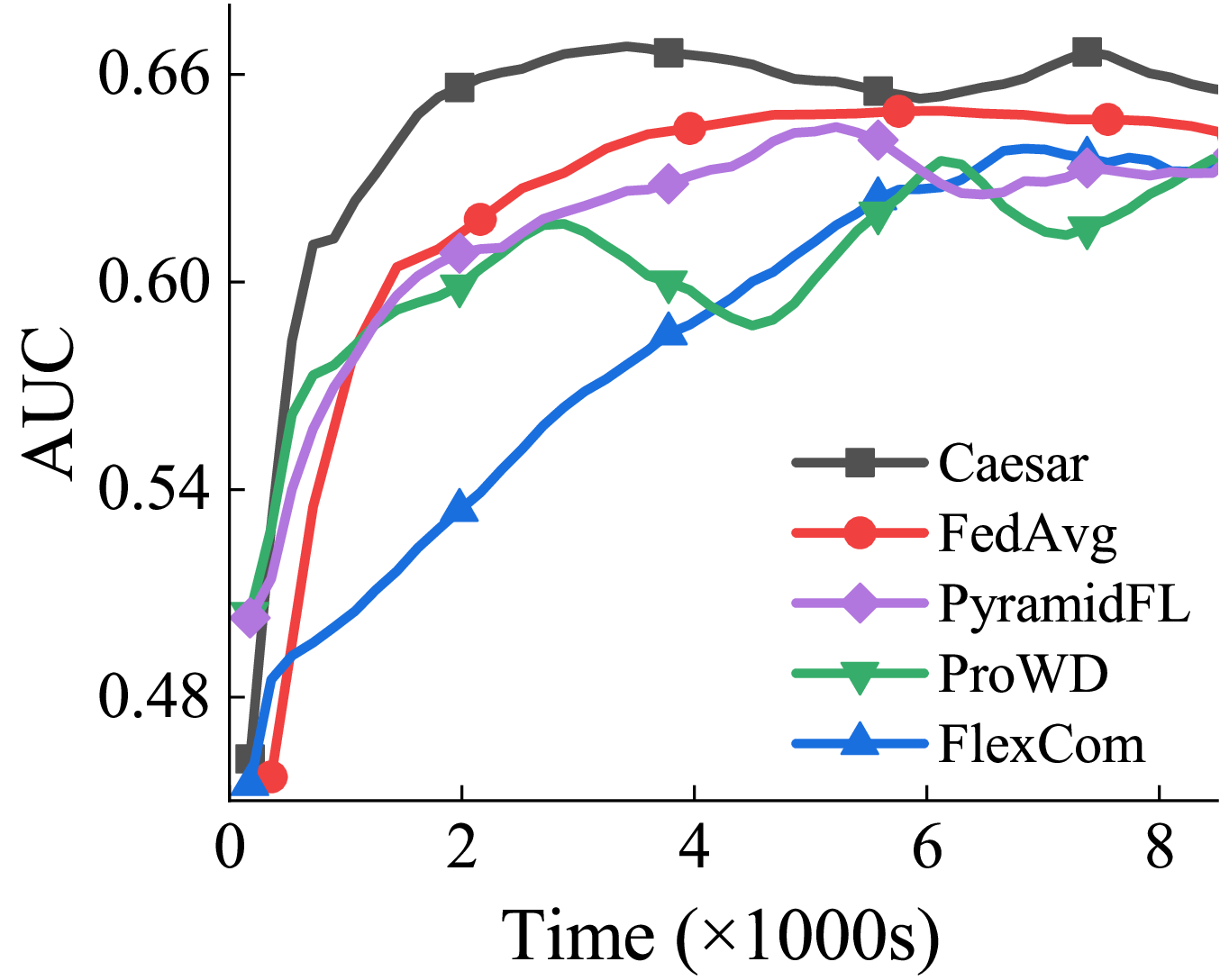}
    \label{fig:oppo_time_acc}
}
\vspace{-3mm}
\caption{Time-to-Accuracy performance of five schemes on the four datasets.}\label{fig:time_acc}
\end{figure*}

\begin{figure*}[t]
\centering

\subfigure[CIFAR-10]
{
    \includegraphics[width=0.22\linewidth,height=3.2cm]{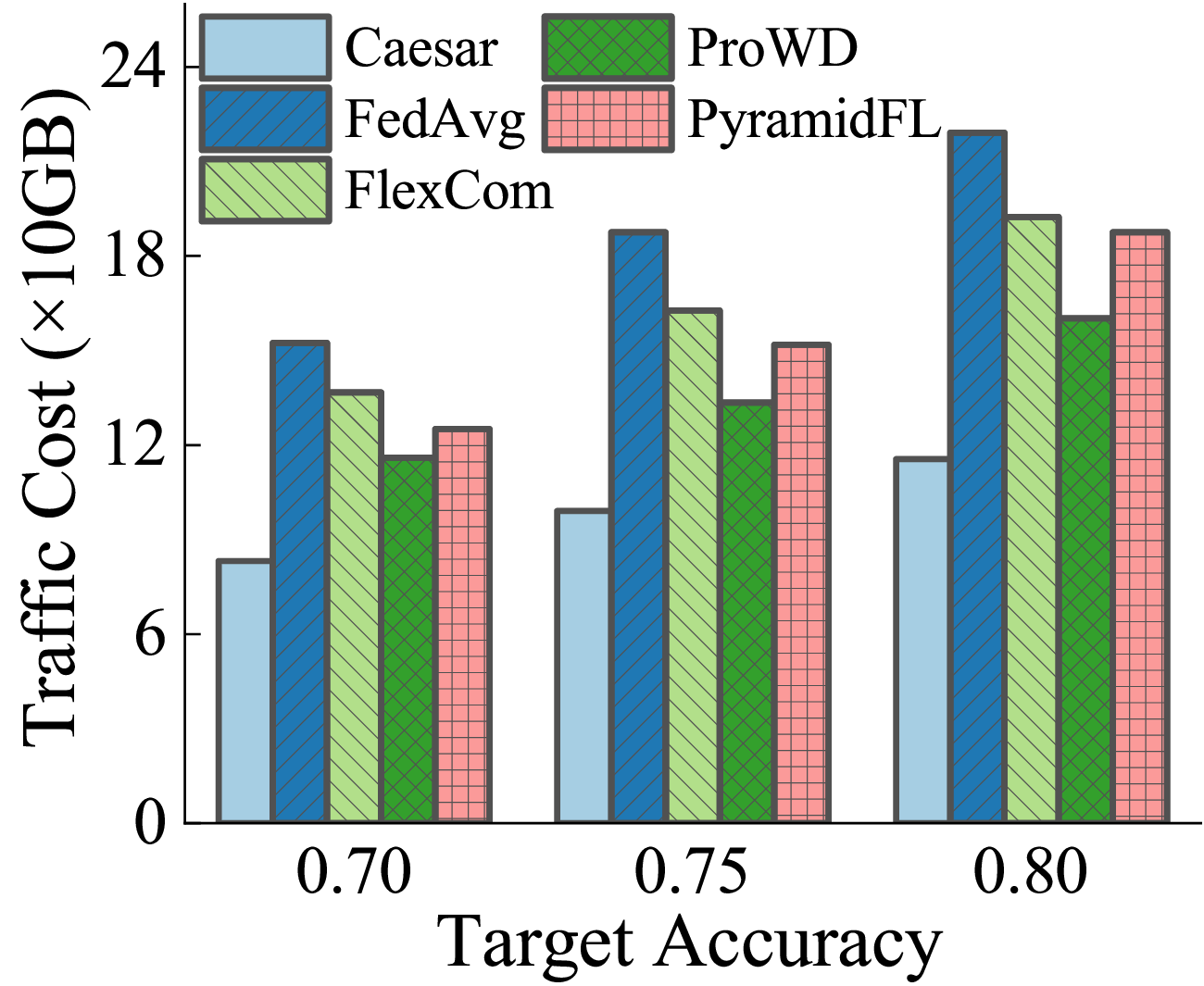}
    \label{fig:cifar_acc_traffic}
}\quad 
\subfigure[HAR]
{
    \includegraphics[width=0.22\linewidth,height=3.2cm]{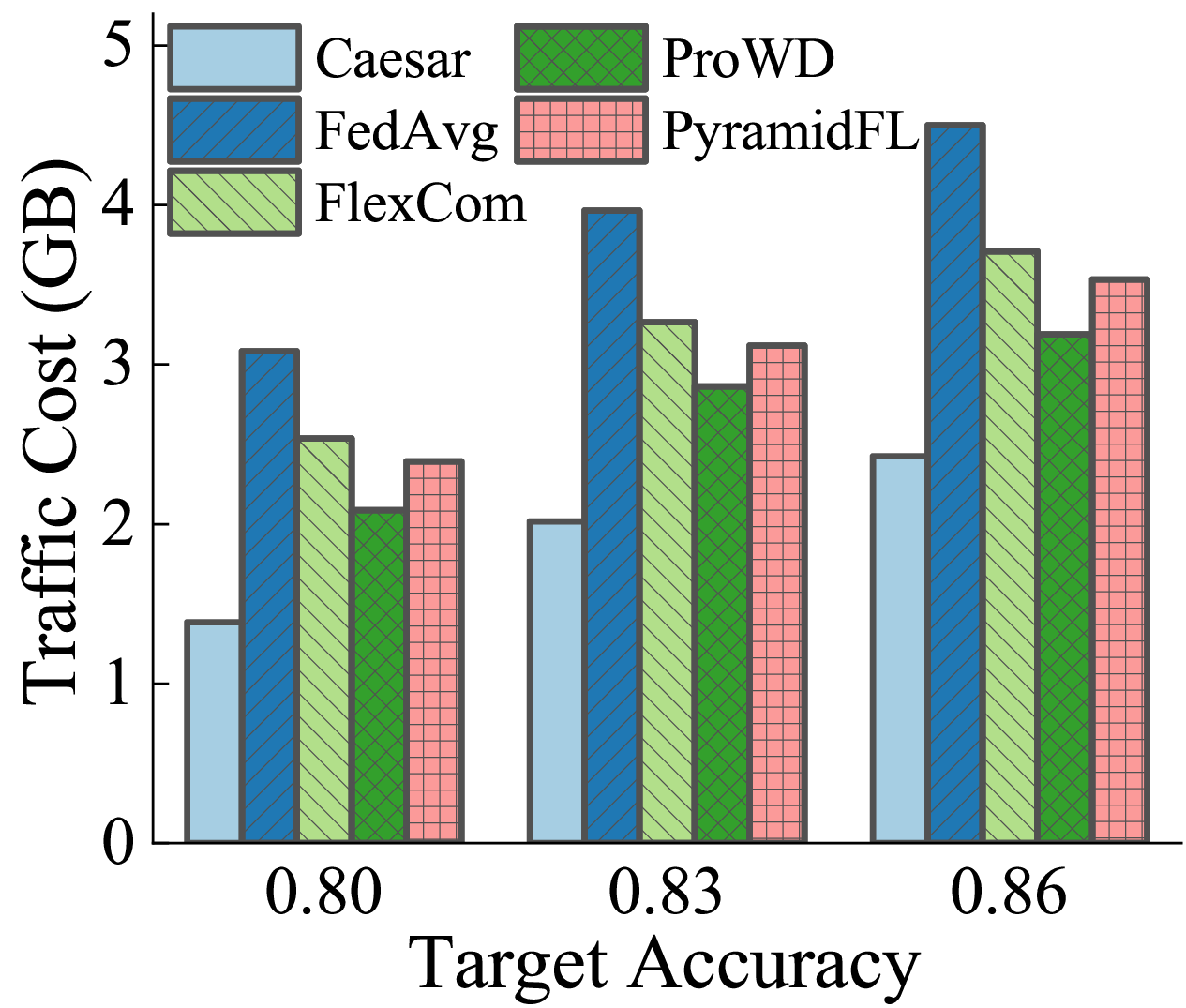}
    \label{fig:har_acc_traffic}
}\quad 
\subfigure[Speech]
{
    \includegraphics[width=0.22\linewidth,height=3.2cm]{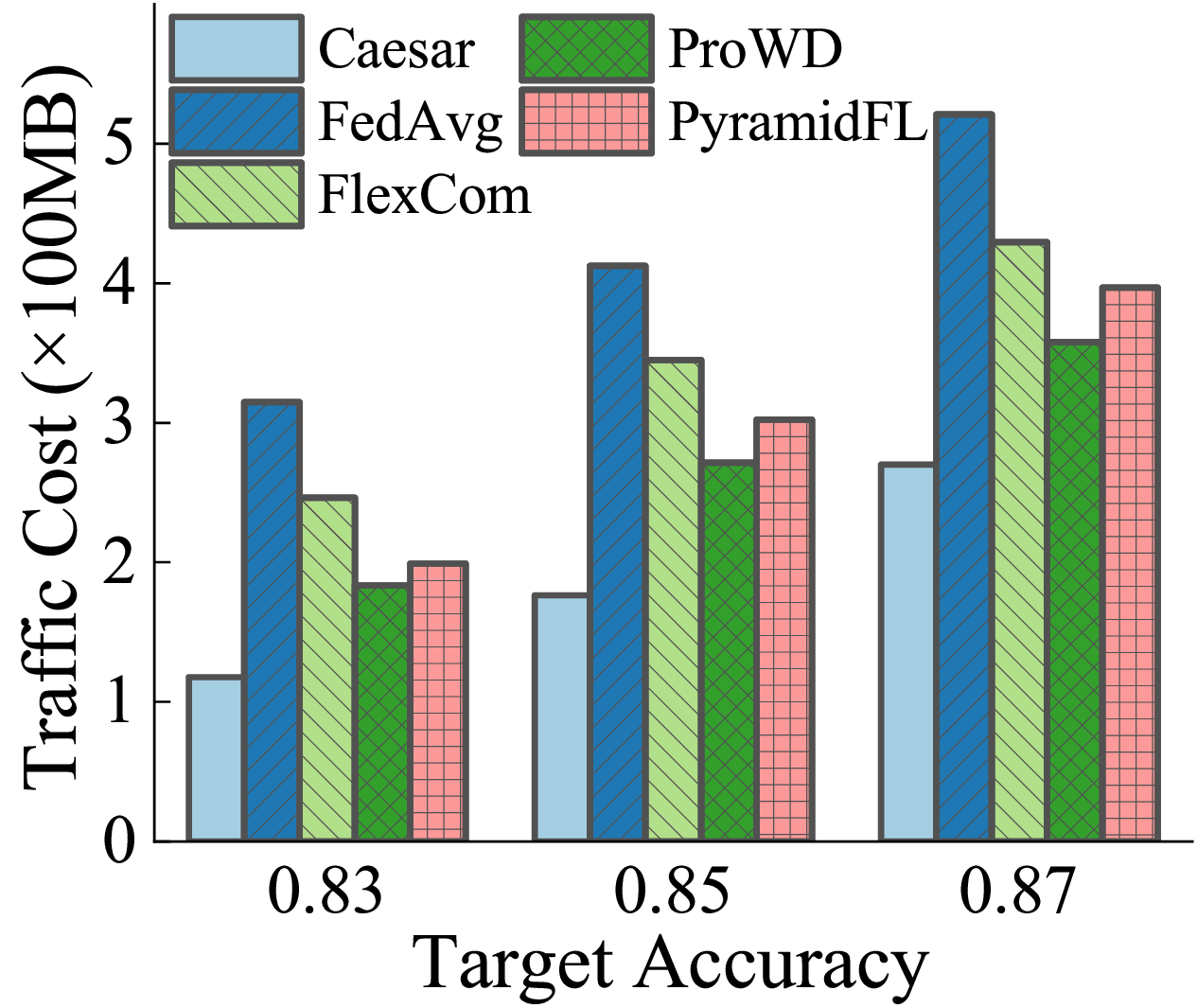}
    \label{fig:speech_acc_traffic}
}\quad 
\subfigure[OPPO-TS]
{
    \includegraphics[width=0.22\linewidth,height=3.2cm]{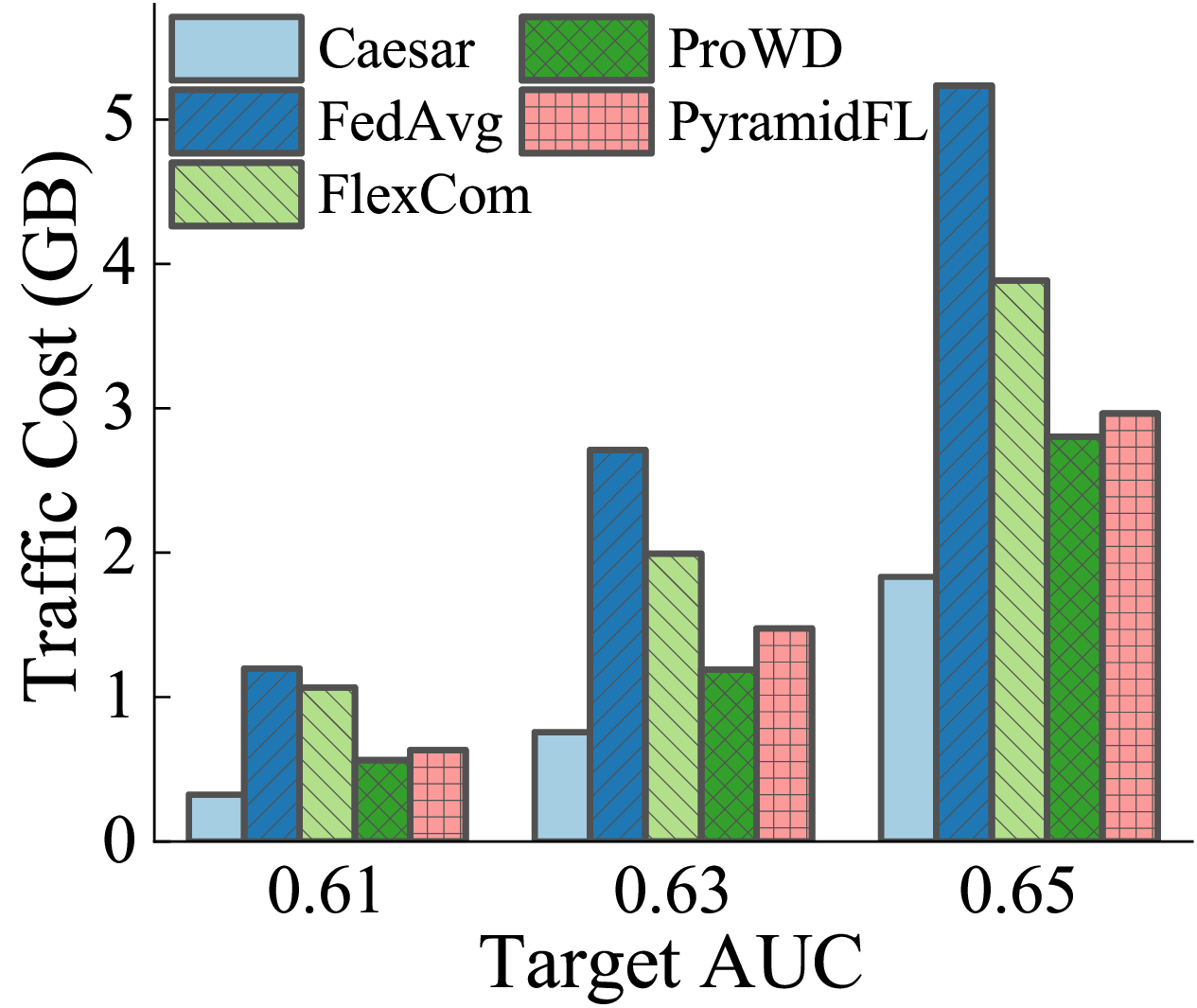}
    \label{fig:oppo_auc_traffic}
}
\vspace{-3mm}
\caption{Traffic-to-Accuracy performance of five schemes on the four datasets.}\label{fig:acc_traffic}
\end{figure*}

\noindent\textbf{Baselines.}
We demonstrate the effectiveness of \method via a comparison with the following four baselines.
\begin{itemize}
    \item \textbf{\fedavg \cite{mcmahan2017communication}} is a fundamental FL scheme, where the transmitted global models and local gradients are uncompressed.
    Devices are configured with an identical and fixed batch size for model training.
    \item \textbf{\flexcom \cite{li2021talk}} compresses the local gradients of participants by Top-K sparsification according to the network condition.
    Besides, the devices adopt an identical and gradually increasing batch size.
    \item \textbf{\prowd \cite{yoon2022bitwidth}} proposes to represent each element in the transmitted model and gradient with fewer bits via quantization.
    The quantization level for each device is determined based on its bandwidth.
    \item \textbf{\pyramidfl \cite{li2022pyramidfl}} ranks the devices by their local gradients norm, and this rank will guide the gradient compression.
    Moreover, it adjusts the number of local iterations on devices to mitigate their waiting time.

\end{itemize}

\noindent\textbf{Evaluation Metrics.}
We adopt the following three metrics to evaluate the performance of \method and baselines.

\textbf{1) Test Accuracy} is measured by the proportion of the samples correctly predicted by the models on the test dataset.
For OPPO-TS, we adopt area under curve (AUC) as an alternative, which is better with a higher value.

\textbf{2) Traffic-to-Accuracy} is the total transmission volume for training the model to achieve the target accuracy. 
Specifically, the traffic costs of both global model download and local gradient upload are recorded.

\textbf{3) Time-to-Accuracy} is defined as the total wall clock time taken to achieve the target accuracy.
In addition, we also record the average waiting time of participants in each round to reflect the impact of the synchronized barrier.

\noindent \textbf{Experimental Parameters.}
By default, the number of communication rounds is configured as 50 for OPPO-TS, 150 for HAR, and 250 for both CIFAR-10 and Speech.
As for the training hyper-parameters, we set the learning rate, decay rate, the number of local iterations, and batch size for HAR as 0.01, 0.98, 10, and 16, respectively.
For the other three datasets, those hyper-parameters are separately set to 0.1, 0.993, 30, and 32 \cite{liao2024mergesfl}.
Regarding the system hyper-parameters, the participate rate $\alpha$ is specified as 0.1 in every experiment, and all five schemes select participants randomly in each communication round for fair comparison.
In schemes that adopt compression, we bound the compression ratio in the range of [0.1, 0.6] \cite{li2022pyramidfl}.

\subsection{Overall Performance}

Firstly, we conduct a set of experiments to evaluate the overall performance of \method and baselines.
By the training processes shown in Figure \ref{fig:time_acc}, the model accuracy of \method always surpasses that of the baselines given the same time cost.
For instance, by Figure \ref{fig:cifar_time_acc}, given a training time budget of 10,000s, \method can achieve a model accuracy of 78.32\% over CIFAR-10, while \fedavg, \flexcom, \prowd, and \pyramidfl can only achieve 62.53\%, 68.07\%, 69.58\%, and 72.77\%, respectively.
In other words, with the constraint of training time, \method can improve the model accuracy by about 5.55\%$\thicksim$15.79\% compared with the baselines.
Besides, \method can accelerate the training process significantly.
For example, by Figure \ref{fig:speech_time_acc}, \method takes about 1,874s to achieve a target model accuracy of 80\% over the Speech dataset, while \fedavg, \flexcom, \prowd, and \pyramidfl separately spend about 5,982s, 4,742s, 4,112s, and 3,651s.
Therefore, \method can provide up to 1.95$\times$$\thicksim$3.19$\times$ speedup for the FL training compared to the baselines with the same target accuracy.
As shown in Figures \ref{fig:har_time_acc} and \ref{fig:oppo_time_acc}, \method also outperforms the baselines in time-to-accuracy performance over the HAR and OPPO-TS datasets.

\begin{figure*}[t]
\centering

\subfigure[CIFAR-10]
{
    \includegraphics[width=0.22\linewidth,height=3.2cm]{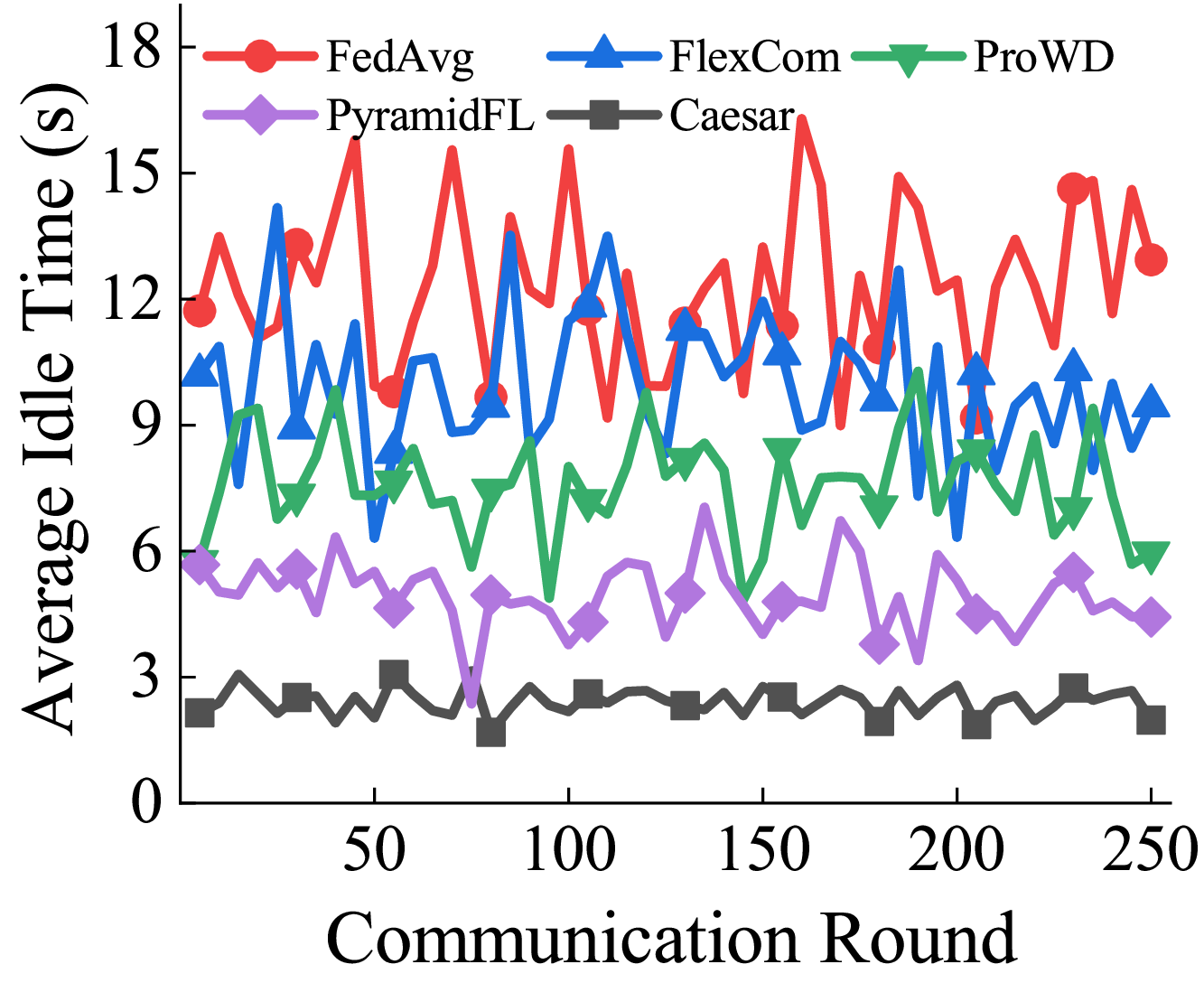}
    \label{fig:cifar_wait}
}\quad 
\subfigure[HAR]
{
    \includegraphics[width=0.22\linewidth,height=3.2cm]{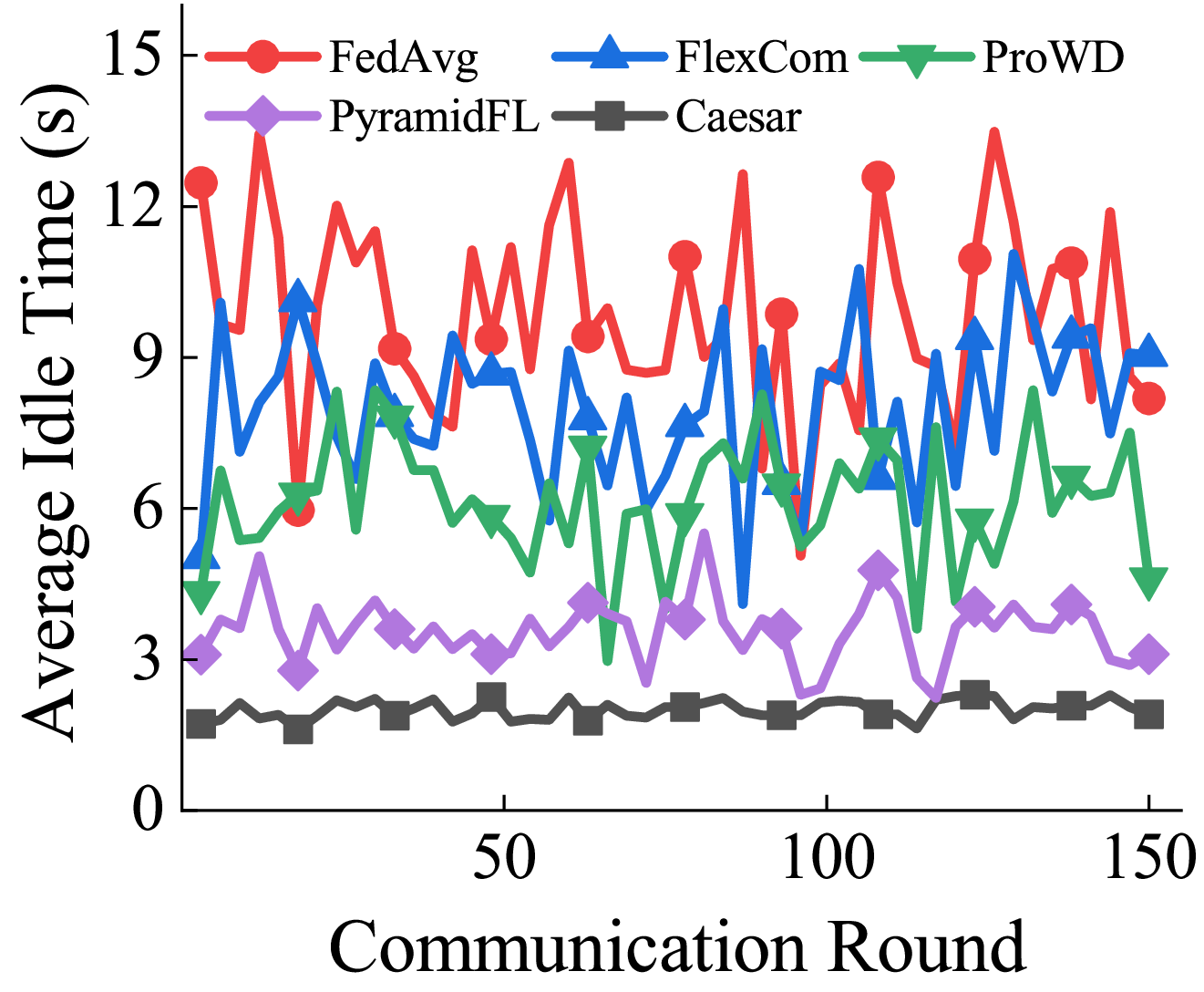}
    \label{fig:har_wait}
}\quad 
\subfigure[Speech]
{
    \includegraphics[width=0.22\linewidth,height=3.2cm]{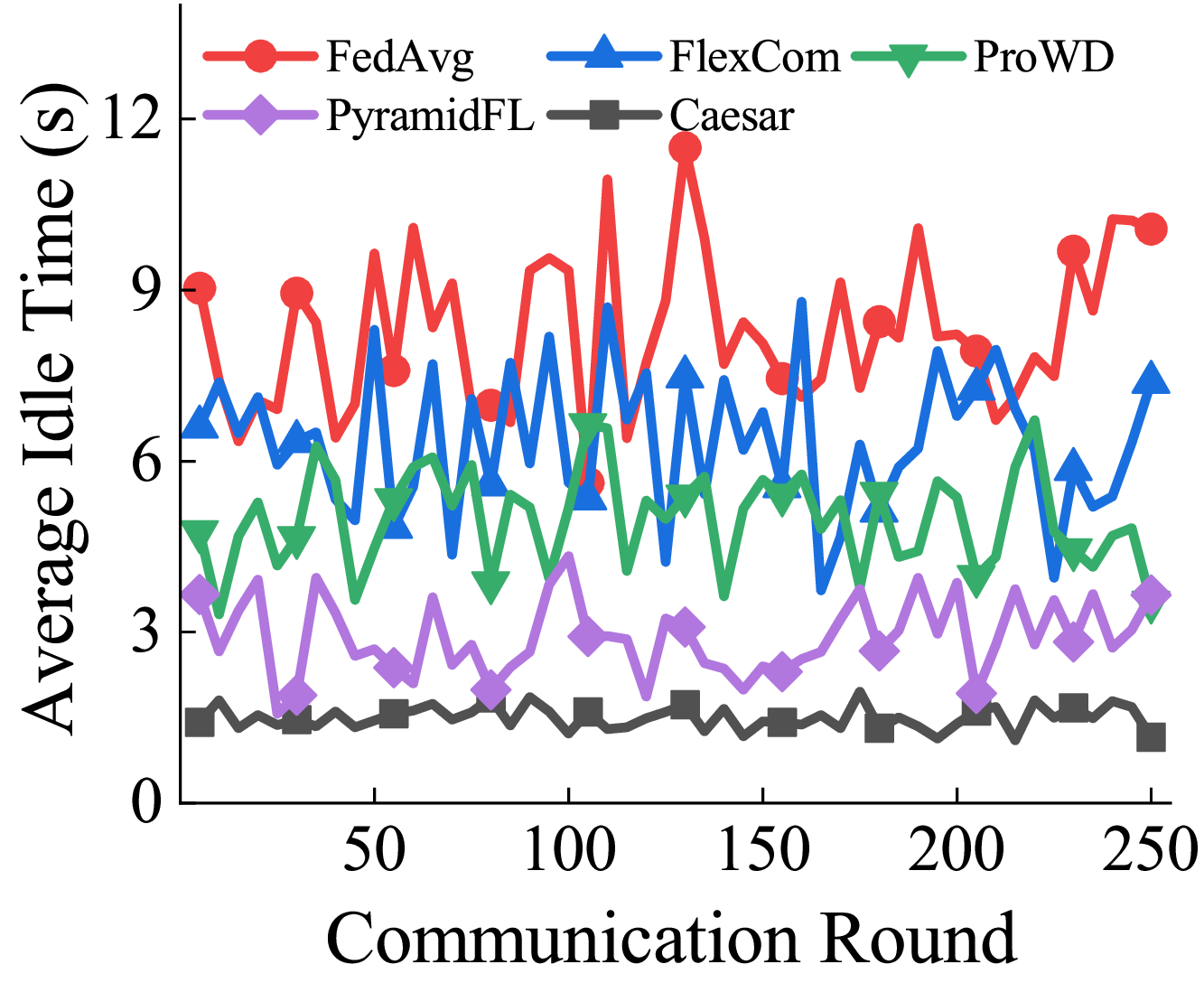}
    \label{fig:speech_wait}
}\quad 
\subfigure[OPPO-TS]
{
    \includegraphics[width=0.22\linewidth,height=3.2cm]{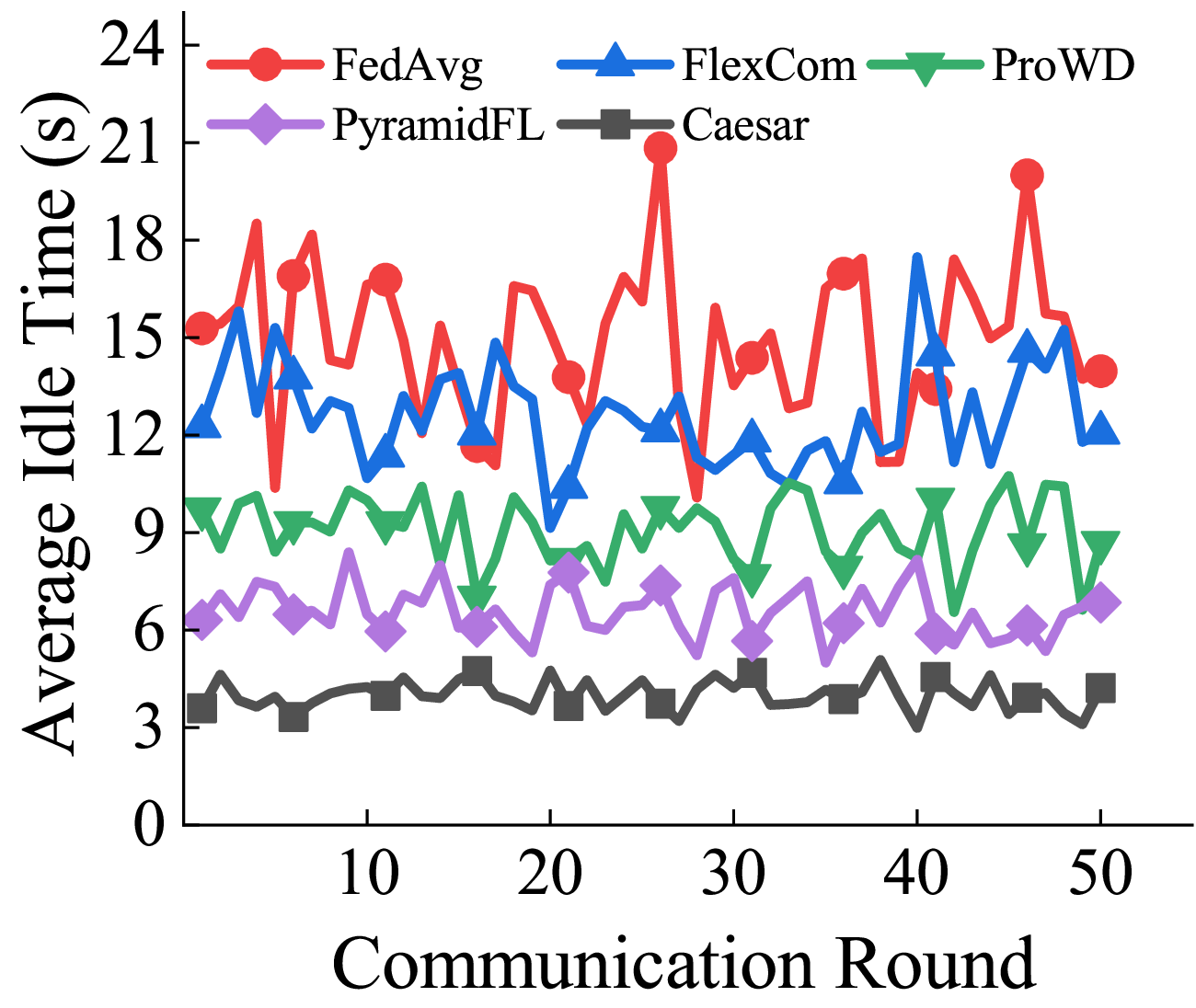}
    \label{fig:oppo_wait}
}
\vspace{-3mm}
\caption{The average waiting time among participants of five schemes on the four datasets.}\label{fig:wait}
\end{figure*}

\begin{figure*}[t]
\centering

\subfigure[CIFAR-10]
{
    \includegraphics[width=0.22\linewidth,height=3.2cm]{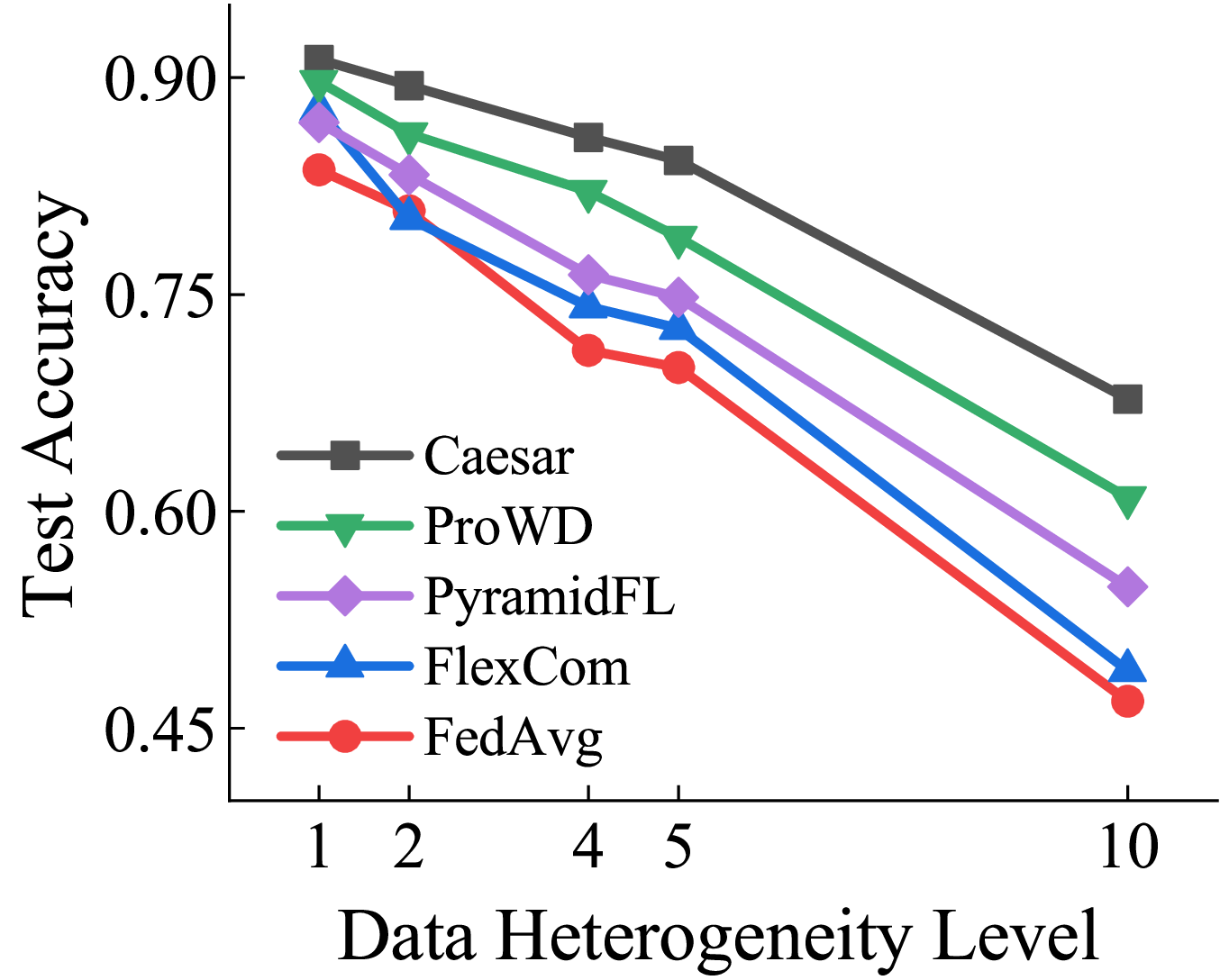}
    \label{fig:cifar_level_acc}
}\quad 
\subfigure[HAR]
{
    \includegraphics[width=0.22\linewidth,height=3.2cm]{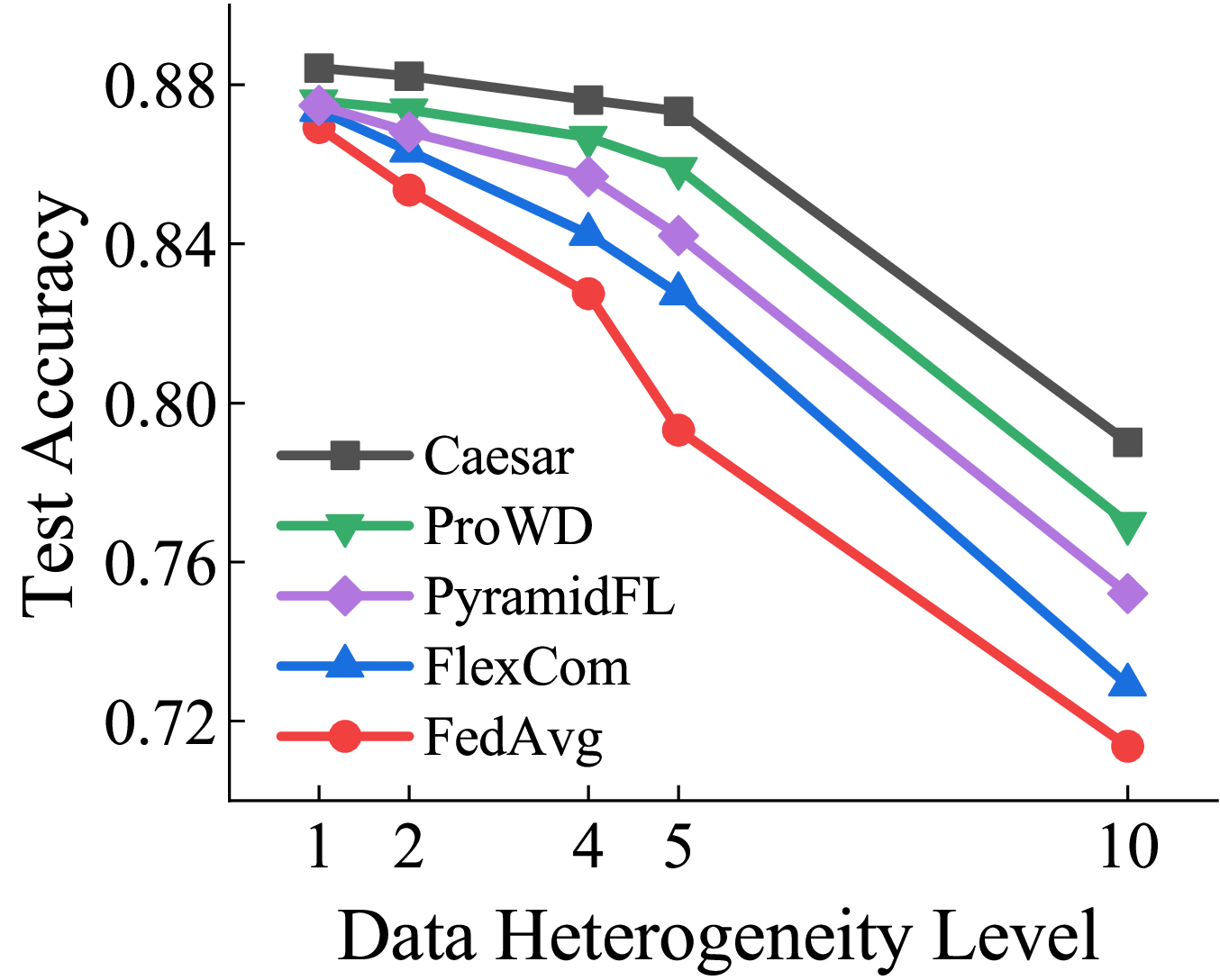}
    \label{fig:har_level_acc}
}\quad 
\subfigure[Speech]
{
    \includegraphics[width=0.22\linewidth,height=3.2cm]{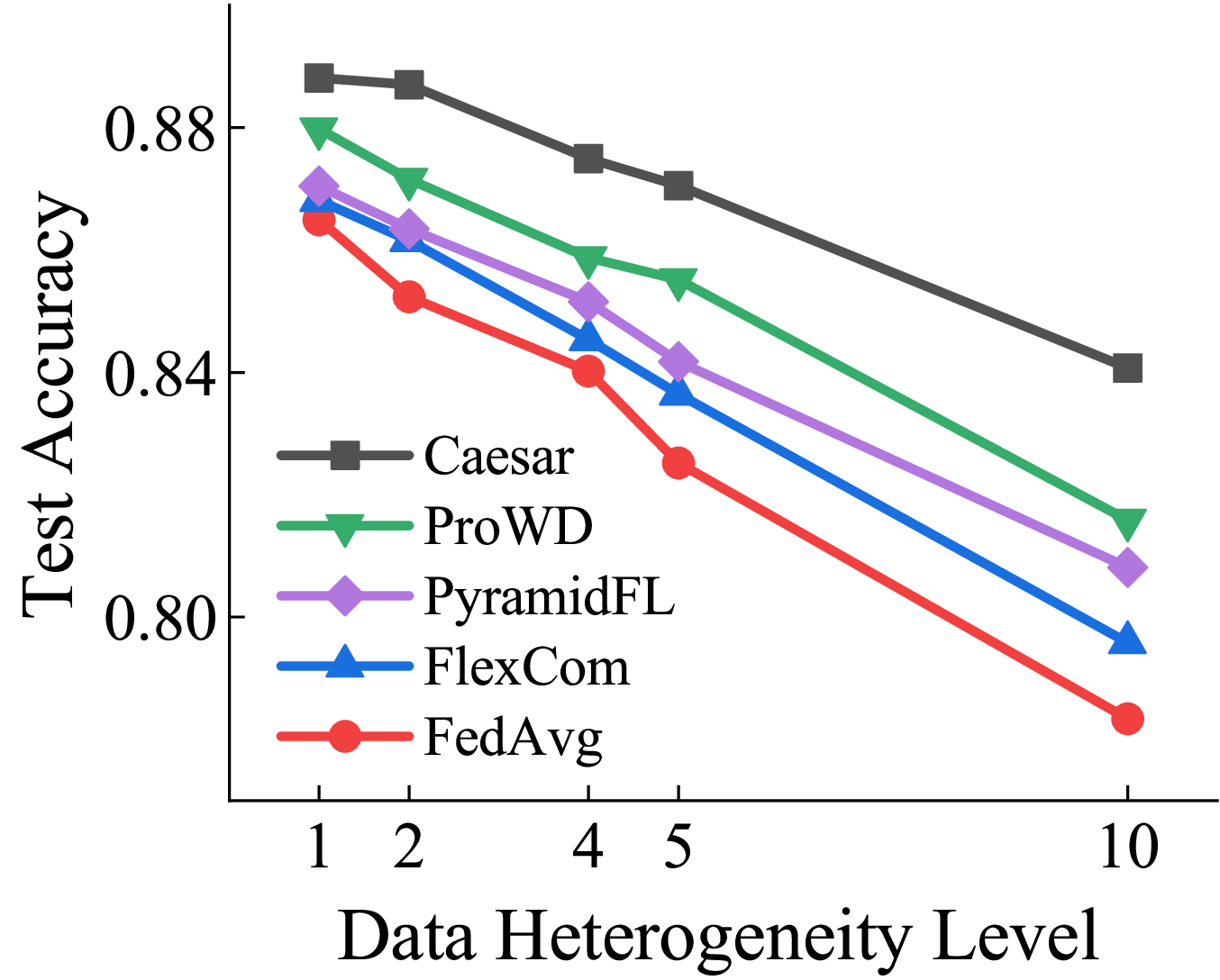}
    \label{fig:speech_level_acc}
}\quad 
\subfigure[Accuracy Degradation]
{
    \includegraphics[width=0.22\linewidth,height=3.2cm]{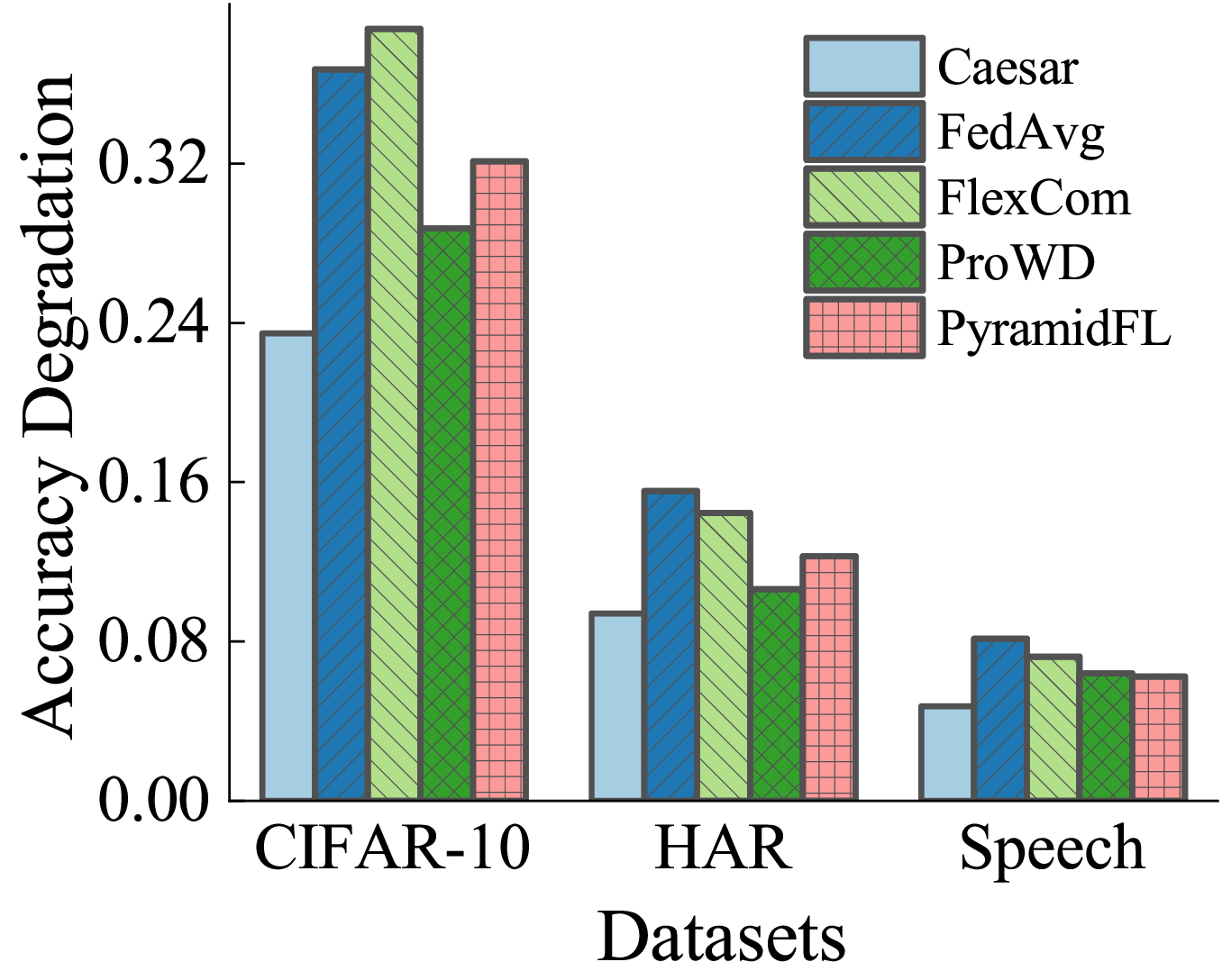}
    \label{fig:dataset_accdelta}
}
\vspace{-3mm}
\caption{The effect of different data heterogeneity levels on five schemes' training performance.}\label{fig:level_acc}
\end{figure*}

Moreover, as shown in Figure \ref{fig:acc_traffic}, \method achieves the best traffic-to-accuracy performance among all schemes.
For instance, by Figure \ref{fig:cifar_acc_traffic}, \method spends 115.57GB network traffic to achieve a target model accuracy of 80\% on the CIFAR-10 dataset, which is about 47.26\%, 39.92\%, 27.87\%, and 38.38\% less than \fedavg, \flexcom, \prowd, and \pyramidfl, respectively.
The similar results are obtained on the other three datasets, which can be found in Figures \ref{fig:har_acc_traffic}$\thicksim$\ref{fig:oppo_auc_traffic}.
We summarize the performance improvement of \method in Table \ref{table:summary}, where the target accuracy or AUC are the highest achievable values for all schemes.
Specifically, Table \ref{table:summary} indicates that \method can mitigate the traffic cost by about 23.94\%$\thicksim$65.06\% and accelerate the FL training by about 1.19$\times$$\thicksim$2.62$\times$ compared to the four baselines on average.
There are two main reasons for such great improvement.
First, the deviation-aware compression strategy can significantly reduce the transmission delay as well as volume in bi-directional communication, while imposing a few impacts on model accuracy.
Next, \method diminishes the idle waiting time on devices via appropriate batch size regulation, which further improves the training efficiency.
Therefore, \method can enjoy better traffic/time-to-accuracy performance than the baselines.

To further demonstrate \method's advantage under the synchronous barrier problem, we illustrate the devices' average waiting time of five schemes.
The results on four datasets impose the same pattern, where \method achieves the least waiting time among all schemes.
Take the CIFAR-10 dataset as an example, the average waiting time of \method is 2.49s, while that of \fedavg, \flexcom, \prowd, and \pyramidfl is 11.68s, 9.76s, 7.41s, and 5.03s, respectively.
\fedavg does not consider the impact of various devices' capabilities, thus resulting in non-negligible waiting time.
\flexcom and \prowd balance the transfer delay of different participants by capability-aware compression, yet the diverse computing latency can still incur significant waiting time.
\pyramidfl can reduce the waiting time to a certain extent via fine-grained adjustment of the number of local iterations on different devices, while the time cost for model download is ignored.

\subsection{Effect of Data Heterogeneity Levels}

Secondly, we observe the impact of data heterogeneity levels on the model training in \method and baselines.
Specifically, we run all five schemes on the CIFAR-10, HAR, and Speech datasets with different data heterogeneity levels (\ie, $p=$ 1, 2, 4, 5, and 10), and compare their final model accuracy, where the traffic budgets for CIFAR-10, HAR, and Speech are 150GB, 30GB, and 300MB, respectively.
The results shown in Figures \ref{fig:cifar_level_acc}$\thicksim$\ref{fig:speech_level_acc} demonstrate that the model accuracy decreases as the data heterogeneity level increases for all schemes.
However, \method consistently outperforms the baselines on all datasets. 
Specifically, by Figure \ref{fig:cifar_level_acc}, with the data heterogeneity level of $p=10$ on CIFAR-10, the model accuracy of \method is 67.78\%, while that of \fedavg, \flexcom, \prowd, and \pyramidfl is only 46.88\%, 48.99\%, 60.95\%, and 54.78\%, respectively.
What's more, by Figure \ref{fig:speech_level_acc}, with a date heterogeneity level of 10, the model accuracy of training with \method is 84.07\% on HAR, which is separately about 5.73\%, 4.49\%, 2.49\%, and 3.26\% higher than that with \fedavg, \flexcom, \prowd, and \pyramidfl.
Moreover, we depict the accuracy degradation of all the five schemes when transforming the level $p$ from 1 to 10 in Figure \ref{fig:dataset_accdelta}.
We can find that \method is more robust to the data heterogeneity.
For instance, due to the intensification of data heterogeneity, \fedavg, \flexcom, \prowd, and \pyramidfl suffer from an accuracy decrease of 8.15\%, 7.23\%, 6.41\%, and 6.23\%, respectively, while \method only loses 4.74\% of model accuracy.
In summary, these results demonstrate that \method can retain advantageous training performance at different data heterogeneity levels, and is more robust compared with the baselines.

\subsection{Effect of Key Strategies}
\begin{figure}[t]
	\centering
        \subfigure[Time-to-Accuracy]{
		\includegraphics[width=1.58in]{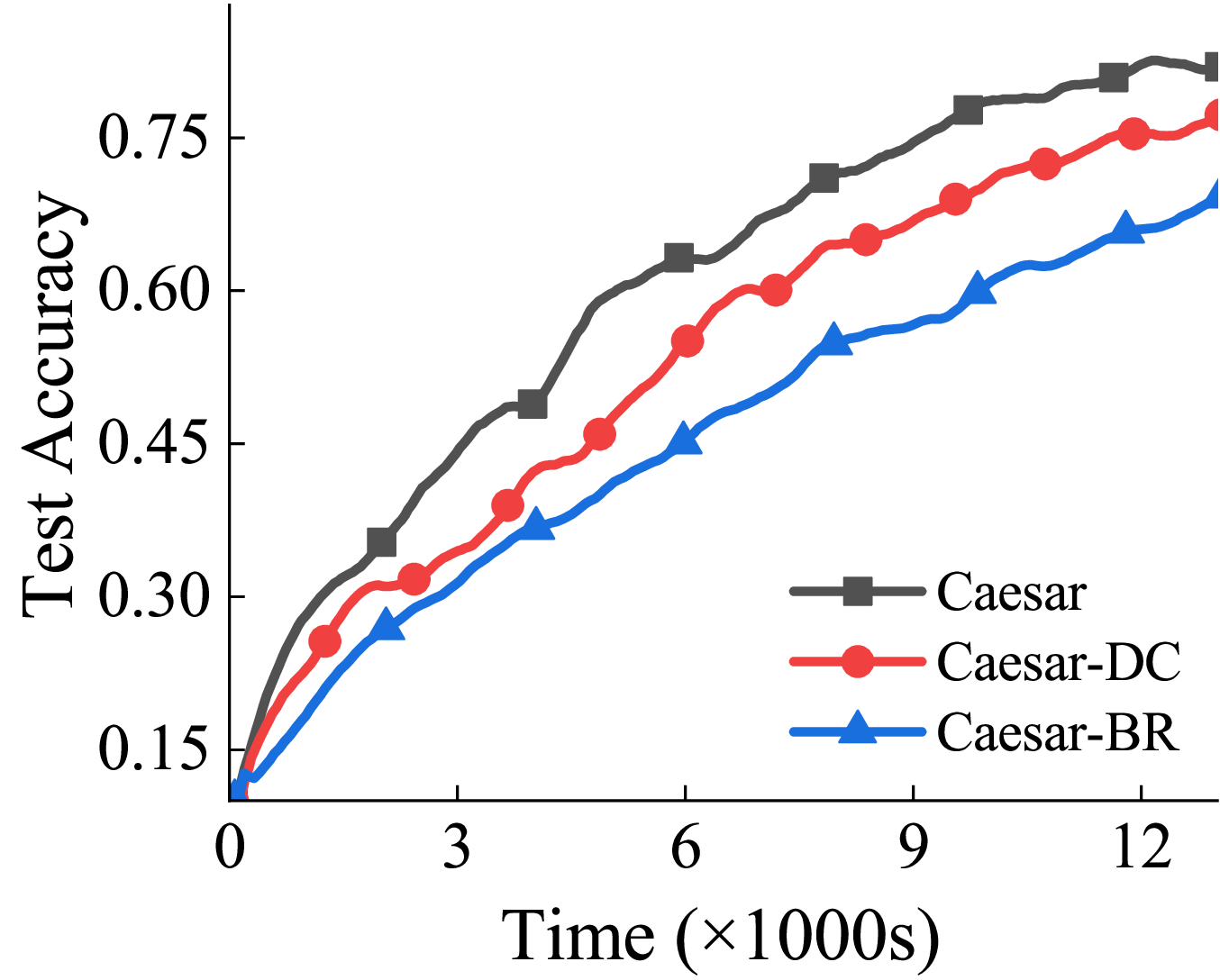}\label{fig:ablation_time_acc}
	}
	\subfigure[Traffic-to-Accuracy]{
		\includegraphics[width=1.58in]{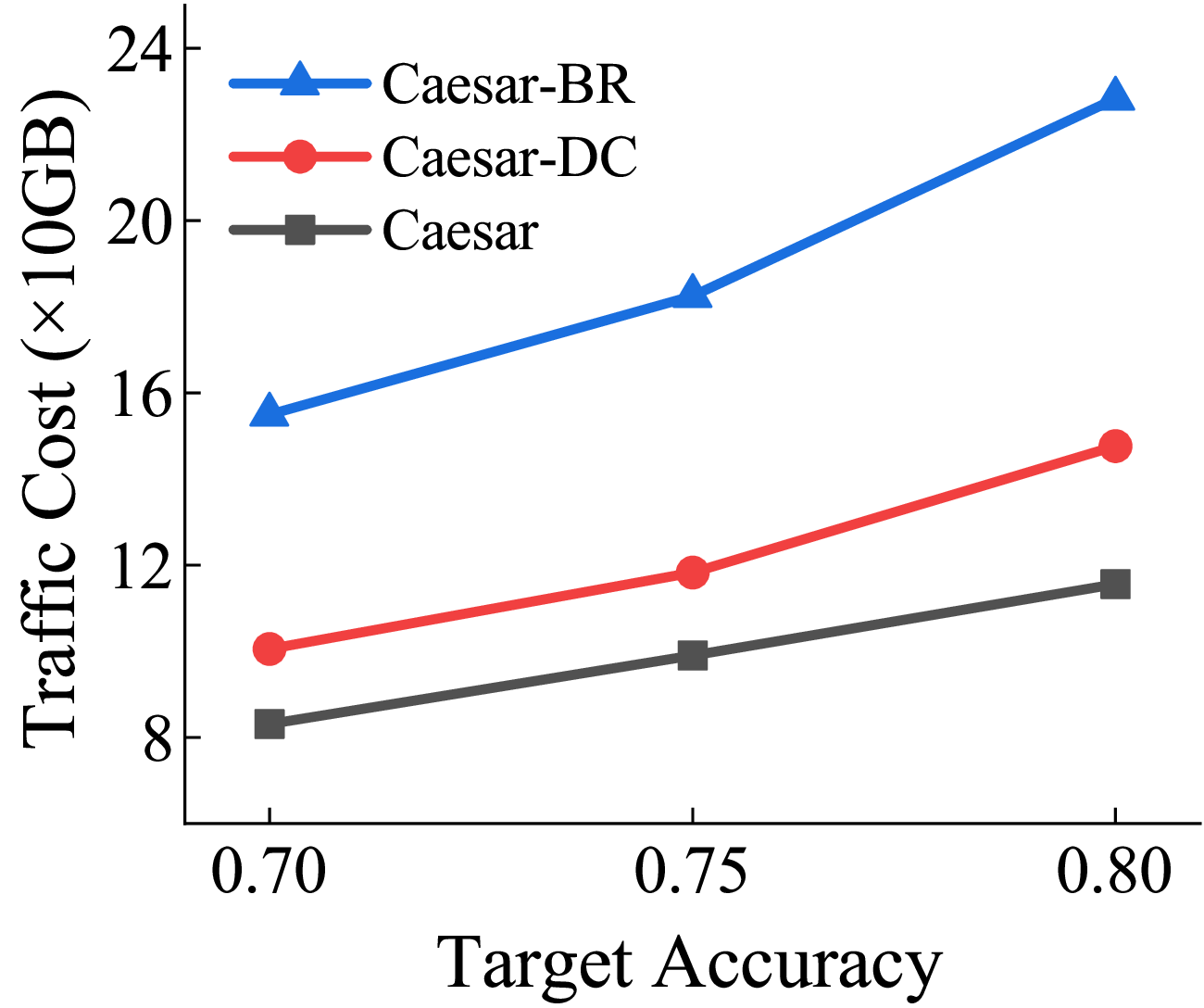}\label{fig:ablation_acc_traffic}
	}
	\caption{\method's ablation study.}\label{fig:ablation}
 \vspace{-3mm}
\end{figure}

Thirdly, we conduct a set of ablation experiments for training ResNet-18 on CIFAR-10 to evaluate the effectiveness of two key strategies in \method, \ie, deviation-aware compression and adaptive batch size regulation.
Specifically, two baselines are adopted for performance comparison.
One is the \method without deviation-aware compression, denoted as \textsf{Caesar-BR}, while another is the \method without adaptive batch size regulation, denoted as \textsf{Caesar-DC}.
The results shown in Figure \ref{fig:ablation} demonstrate that \method's performance declines when either strategy is disabled.
For instance, as illustrated in Figure \ref{fig:ablation_time_acc}, \method takes 8,615s to achieve a target accuracy of 80\% on CIFAR-10, while \textsf{Caesar-BR} and \textsf{Caesar-DC} require 12,043s and 17,818s, respectively.
Besides, by Figure \ref{fig:ablation_acc_traffic}, with the same target accuracy of 80\%, the traffic cost of \method is 115.57GB, while that of \textsf{Caesar-BR} and \textsf{Caesar-DC} is separately 228.31GB and 147.59GB.
Thus, the adaptive batch size regulation can speed up the training by about 2.09$\times$ and reduce the traffic cost by about 21.69\%.
In other words, the strategy of adaptive batch size regulation contributes to a 1.39$\times$ training speedup and a 21.69\% traffic reduction.
Besides, powered by deviation-aware compression, \method can accelerate the training process by about 2.07$\times$, and save the traffic cost by about 49.38\%.
Therefore, both the two key strategies are necessary and important, while the deviation-aware compression plays a more critical role in \method's performance.

\subsection{Effect of Device Scales}
\begin{figure}[t]
	\centering
        \subfigure[Time-to-Accuracy]{
		\includegraphics[width=1.58in]{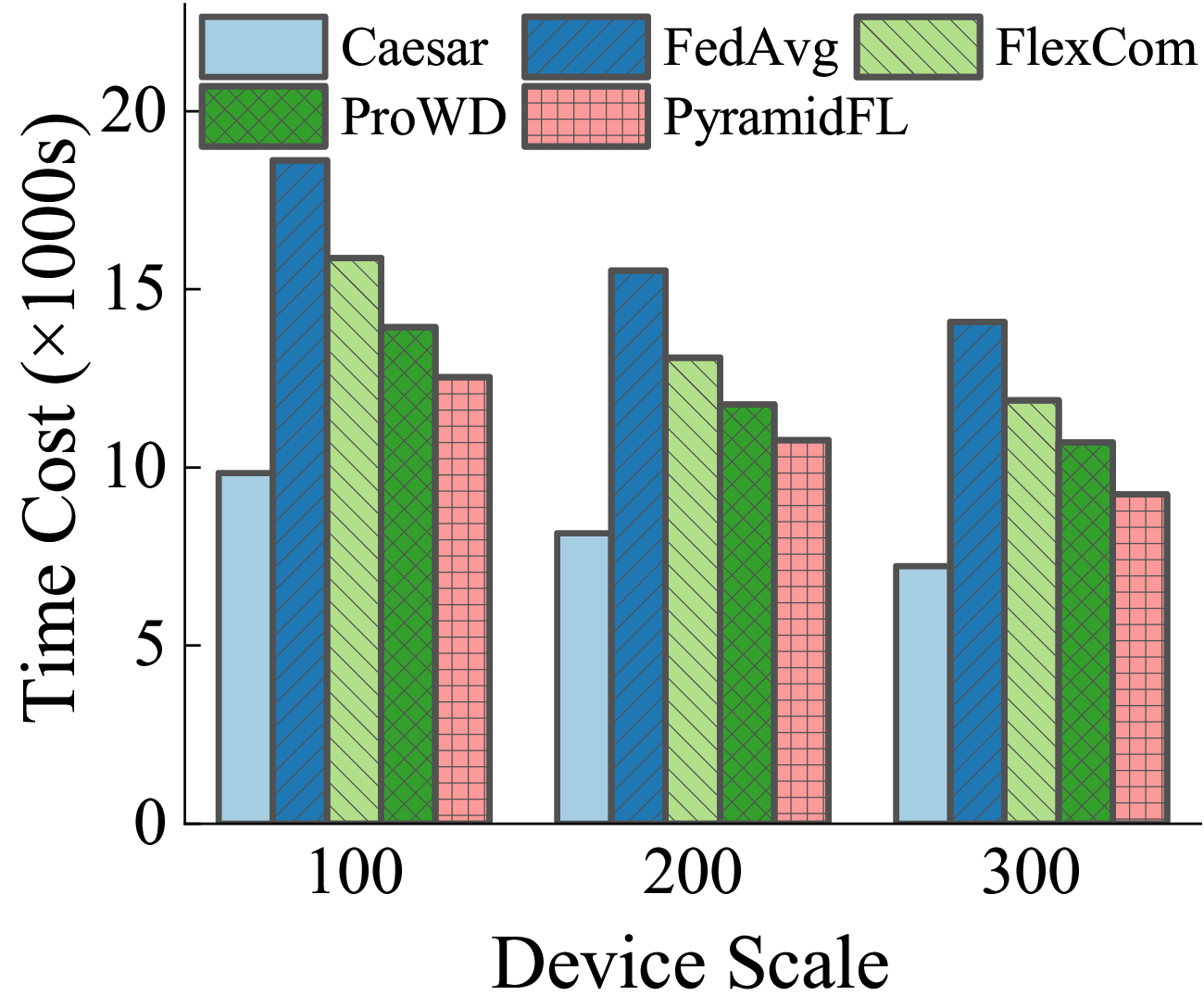}\label{fig:scale_acc_time}
	}
        \subfigure[Traffic-to-Accuracy]{
		\includegraphics[width=1.58in]{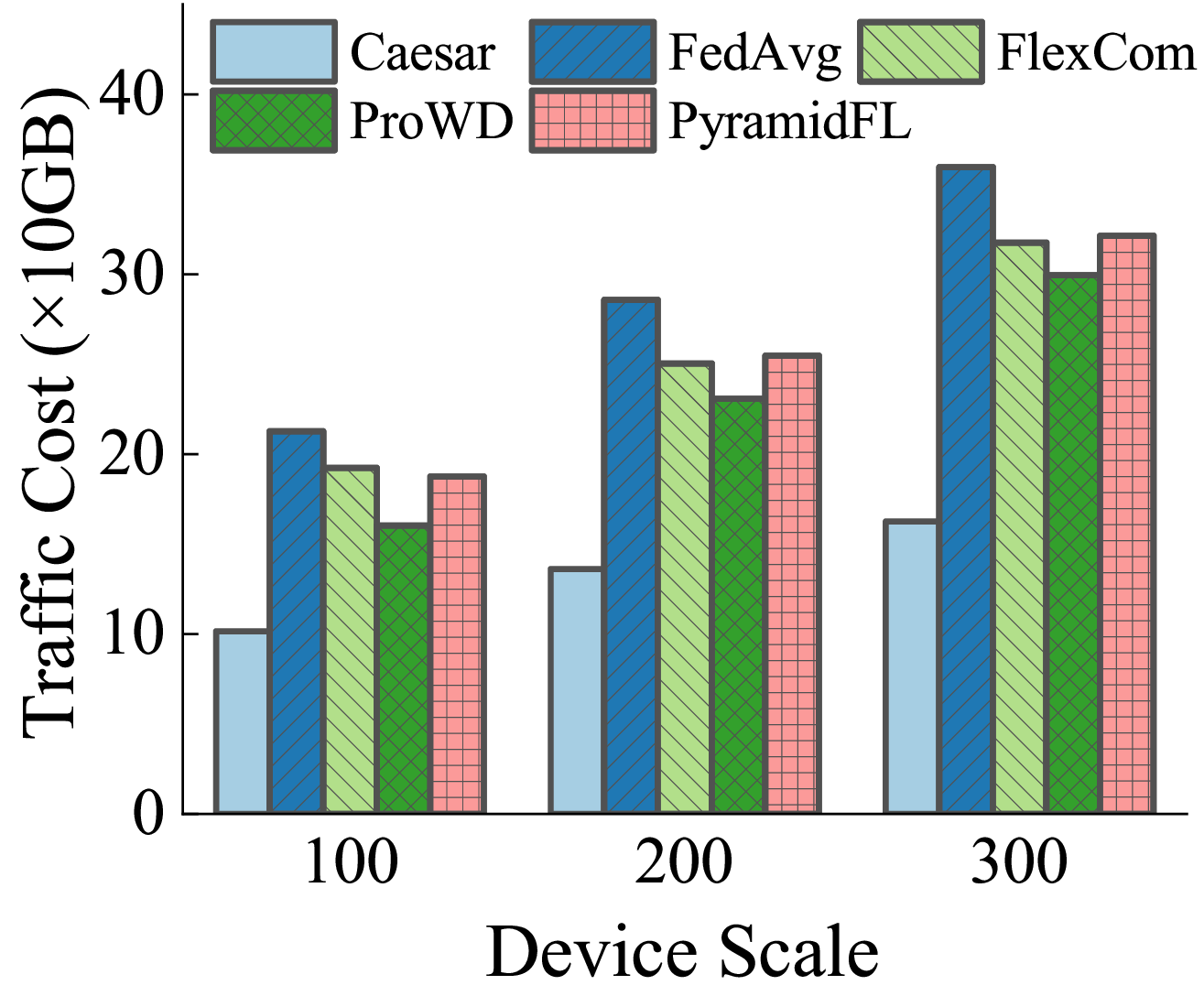}\label{fig:scale_acc_traffic}
	}
	\caption{Performance under different device scales.}\label{fig:scale}
 \vspace{-3mm}
\end{figure}

Finally, we evaluate the scalability and robustness of \method and baselines with three different device scales (\ie, 100, 200, and 300) on CIFAR-10.
Due to the limited scale of the physical prototype systems, this set of experiments is conducted in an AMAX deep learning workstation equipped with an Intel(R) Xeon(R) Platinum 8358P CPU, 512GB RAM, and 8 NVIDIA RTX A6000 GPUs (48GB), where each device is simulated as a process of Linux.
The target accuracy is set to 80\% in this subsection.as shown in Figure \ref{fig:scale_acc_time}, the time costs of these five schemes decrease as the device scale increase.
This is because more data samples are utilized for model training in each communication round, resulting in a faster convergence rate.
Under different devices scales, \method is always the fastest among all schemes.
Specifically, \method can achieve a speedup of 1.29$\times$$\thicksim$1.91$\times$ for FL training compared with the four baselines on average.
However, by Figure \ref{fig:scale_acc_traffic}, larger device scale will also incur more traffic cost, but \method always achieves better traffic-to-accuracy performance than the four baselines.
On average, \method can reduce the traffic cost by about 41.01\%$\thicksim$53.04\%  compared to the baselines.
These results further demonstrate the excellent scalability and robustness of \method, indicating its high potential for application in large-scale systems.


\section{Related Work}\label{sec:related}
FL is a practical and promising scheme for privacy-preserving model training, which has garnered significant interest from both research and industrial communities.
However, the real-world FL system often suffers from the contradiction between high model accuracy and light traffic cost \cite{nguyen2021federated}.
In recent years, a lot of efforts have been made to relieve the communication overhead for FL \cite{zhao2023towards}.
We summarize the related works into the following two categories:

\textbf{Compression-based Schemes. }
It is a common and natural solution to compress the transmitted models or gradients by various techniques such as quantization \cite{liu2023communication, cui2022optimal, honig2022dadaquant, bernstein2018signsgd, alistarh2017qsgd}, sparsification \cite{xu2021deepreduce, sattler2019robust, horvath2021fjord, diao2020heterofl, sattler2019robust}, and tensor decomposition \cite{mei2022resource, wang2023svdfed, hyeon2021fedpara}.
For example, \textsf{QSGD} \cite{alistarh2017qsgd} represents each model parameter with fewer bits via quantization.
\textsf{Hermes} \cite{li2021hermes} reduces the model size by pruning the unimportant parameters.
\textsf{SVDFed} \cite{wang2023svdfed} proposes to utilize singular value decomposition to decompose the gradient into lightweight low-rank tensors.
While these works can efficiently relieve the payload within every communication round, the compression often causes excessive deviation for the local training and/or the global aggregation, which will slow the model convergence rate or even decrease the final test accuracy, especially under the challenges of data heterogeneity and model obsolescence.
To tackle this issue, we propose a novel deviation-aware compression approach, which can save the traffic cost significantly while minimizing the impact of model/gradient deviation on the training performance.

\textbf{Speedup-based Schemes. }
Another line of works \cite{wang2018edge, lai2021oort, li2022pyramidfl, wang2020tackling, ye2023feddisco, luo2022tackling} aim to accelerate the convergence for FL, so that the model will require fewer communication rounds (\ie, less traffic cost) to achieve the target accuracy.
For example, authors in \cite{wang2018edge} adjust the local update frequency to obtain a tighter theoretical upper bound, which means a faster convergence rate.
\textsf{Oort} \cite{lai2021oort} and \textsf{PyramidFL} \cite{li2022pyramidfl} select a list of participants with high statistical utility to participate in each communication round, showing the superior time-to-accuracy performance over the random selection strategy.
\textsf{FedDisco} \cite{ye2023feddisco} optimizes the aggregation weights of participants by taking both the dataset size and the discrepancy into consideration, so as to accelerate the convergence.
Different from these speedup-based schemes, we mainly focus on optimizing the compression approach to achieve a delicate trade-off between traffic cost and model accuracy.
Meanwhile, We impose no limitation on the local update frequency configuration, device selection strategy, or aggregation weight algorithm, making \method compatible with these speedup-based schemes.

\section{Conclusion}\label{sec:conclusion} 
In this work, we propose \method, an efficient FL system with a novel deviation-aware compression approach.
Specifically, to address the challenge of model obsolescence, \method cherry-picks the downstream compression ratios according to devices' staleness, providing a precise initial model for every participant's local training.
Besides, to ensure that global model can learn important knowledge from the heterogeneous data, \method estimates each device' importance based on its data properties to guide the determination of upstream compression ratio.
Moreover, \method also employs fine-grained batch size optimization to reduce devices' waiting times under the synchronous barrier, thereby accelerating the training process.
We have implemented \method on two physical platforms for the performance evaluation. 
Extensive results demonstrate the advantages of \method compared to the existing FL systems.

\bibliographystyle{unsrt}
\bibliography{main}

\begin{thebibliography}{10}

\bibitem{niu2020billion}
Chaoyue Niu, Fan Wu, Shaojie Tang, Lifeng Hua, Rongfei Jia, Chengfei Lv, Zhihua Wu, and Guihai Chen.
\newblock Billion-scale federated learning on mobile clients: A submodel design with tunable privacy.
\newblock In {\em Proceedings of the 26th Annual International Conference on Mobile Computing and Networking}, pages 1--14, 2020.

\bibitem{zheng2023autofed}
Tianyue Zheng, Ang Li, Zhe Chen, Hongbo Wang, and Jun Luo.
\newblock Autofed: Heterogeneity-aware federated multimodal learning for robust autonomous driving.
\newblock In {\em Proceedings of the 29th Annual International Conference on Mobile Computing and Networking}, pages 1--15, 2023.

\bibitem{enwiki:1208869058}
{Wikipedia contributors}.
\newblock General data protection regulation --- {Wikipedia}{,} the free encyclopedia, 2024.
\newblock [Online; accessed 20-February-2024].

\bibitem{enwiki:1198023421}
{Wikipedia contributors}.
\newblock California consumer privacy act --- {Wikipedia}{,} the free encyclopedia, 2024.
\newblock [Online; accessed 20-February-2024].

\bibitem{mcmahan2017communication}
Brendan McMahan, Eider Moore, Daniel Ramage, Seth Hampson, and Blaise~Aguera y~Arcas.
\newblock Communication-efficient learning of deep networks from decentralized data.
\newblock In {\em Artificial intelligence and statistics}, pages 1273--1282. PMLR, 2017.

\bibitem{liu2023yoga}
Jun Liu, Jianchun Liu, Hongli Xu, Yunming Liao, Zhiyuan Wang, and Qianpiao Ma.
\newblock Yoga: Adaptive layer-wise model aggregation for decentralized federated learning.
\newblock {\em IEEE/ACM Transactions on Networking}, 2023.

\bibitem{liu2023adaptive}
Jianchun Liu, Qingmin Zeng, Hongli Xu, Yang Xu, Zhiyuan Wang, and He~Huang.
\newblock Adaptive block-wise regularization and knowledge distillation for enhancing federated learning.
\newblock {\em IEEE/ACM Transactions on Networking}, 2023.

\bibitem{liu2021adaptive}
Jianchun Liu, Hongli Xu, Lun Wang, Yang Xu, Chen Qian, Jinyang Huang, and He~Huang.
\newblock Adaptive asynchronous federated learning in resource-constrained edge computing.
\newblock {\em IEEE Transactions on Mobile Computing}, 22(2):674--690, 2021.

\bibitem{jiang2022fedmp}
Zhida Jiang, Yang Xu, Hongli Xu, Zhiyuan Wang, Chunming Qiao, and Yangming Zhao.
\newblock Fedmp: Federated learning through adaptive model pruning in heterogeneous edge computing.
\newblock In {\em 2022 IEEE 38th International Conference on Data Engineering (ICDE)}, pages 767--779. IEEE, 2022.

\bibitem{kenton2019bert}
Jacob Devlin Ming-Wei~Chang Kenton and Lee~Kristina Toutanova.
\newblock Bert: Pre-training of deep bidirectional transformers for language understanding.
\newblock In {\em Proceedings of naacL-HLT}, volume~1, page~2. Minneapolis, Minnesota, 2019.

\bibitem{lang1995newsweeder}
Ken Lang.
\newblock Newsweeder: Learning to filter netnews.
\newblock In {\em Machine learning proceedings 1995}, pages 331--339. Elsevier, 1995.

\bibitem{lin2021fednlp}
Bill~Yuchen Lin, Chaoyang He, Zihang Zeng, Hulin Wang, Yufen Huang, Christophe Dupuy, Rahul Gupta, Mahdi Soltanolkotabi, Xiang Ren, and Salman Avestimehr.
\newblock Fednlp: Benchmarking federated learning methods for natural language processing tasks.
\newblock {\em arXiv preprint arXiv:2104.08815}, 2021.

\bibitem{cai2023efficient}
Dongqi Cai, Yaozong Wu, Shangguang Wang, Felix~Xiaozhu Lin, and Mengwei Xu.
\newblock Efficient federated learning for modern nlp.
\newblock In {\em Proceedings of the 29th Annual International Conference on Mobile Computing and Networking}, pages 1--16, 2023.

\bibitem{nguyen2021federated}
Dinh~C Nguyen, Ming Ding, Quoc-Viet Pham, Pubudu~N Pathirana, Long~Bao Le, Aruna Seneviratne, Jun Li, Dusit Niyato, and H~Vincent Poor.
\newblock Federated learning meets blockchain in edge computing: Opportunities and challenges.
\newblock {\em IEEE Internet of Things Journal}, 8(16):12806--12825, 2021.

\bibitem{kairouz2021advances}
Peter Kairouz, H~Brendan McMahan, Brendan Avent, Aur{\'e}lien Bellet, Mehdi Bennis, Arjun~Nitin Bhagoji, Kallista Bonawitz, Zachary Charles, Graham Cormode, Rachel Cummings, et~al.
\newblock Advances and open problems in federated learning.
\newblock {\em Foundations and trends{\textregistered} in machine learning}, 14(1--2):1--210, 2021.

\bibitem{jiang2023heterogeneity}
Zhida Jiang, Yang Xu, Hongli Xu, Zhiyuan Wang, and Chen Qian.
\newblock Heterogeneity-aware federated learning with adaptive client selection and gradient compression.
\newblock In {\em IEEE INFOCOM 2023-IEEE Conference on Computer Communications}, pages 1--10. IEEE, 2023.

\bibitem{wang2023accelerating}
Lun Wang, Yang Xu, Hongli Xu, Min Chen, and Liusheng Huang.
\newblock Accelerating decentralized federated learning in heterogeneous edge computing.
\newblock {\em IEEE Transactions on Mobile Computing}, 22(9):5001--5016, 2023.

\bibitem{wang2020optimizing}
Hao Wang, Zakhary Kaplan, Di~Niu, and Baochun Li.
\newblock Optimizing federated learning on non-iid data with reinforcement learning.
\newblock In {\em IEEE INFOCOM 2020-IEEE conference on computer communications}, pages 1698--1707. IEEE, 2020.

\bibitem{liu2022enhancing}
Jianchun Liu, Yang Xu, Hongli Xu, Yunming Liao, Zhiyuan Wang, and He~Huang.
\newblock Enhancing federated learning with intelligent model migration in heterogeneous edge computing.
\newblock In {\em 2022 IEEE 38th International Conference on Data Engineering (ICDE)}, pages 1586--1597. IEEE, 2022.

\bibitem{bonawitz2019towards}
Keith Bonawitz, Hubert Eichner, Wolfgang Grieskamp, Dzmitry Huba, Alex Ingerman, Vladimir Ivanov, Chloe Kiddon, Jakub Kone{\v{c}}n{\`y}, Stefano Mazzocchi, Brendan McMahan, et~al.
\newblock Towards federated learning at scale: System design.
\newblock {\em Proceedings of machine learning and systems}, 1:374--388, 2019.

\bibitem{cui2022optimal}
Laizhong Cui, Xiaoxin Su, Yipeng Zhou, and Jiangchuan Liu.
\newblock Optimal rate adaption in federated learning with compressed communications.
\newblock In {\em IEEE INFOCOM 2022-IEEE Conference on Computer Communications}, pages 1459--1468. IEEE, 2022.

\bibitem{xu2021deepreduce}
Hang Xu, Kelly Kostopoulou, Aritra Dutta, Xin Li, Alexandros Ntoulas, and Panos Kalnis.
\newblock Deepreduce: A sparse-tensor communication framework for federated deep learning.
\newblock {\em Advances in Neural Information Processing Systems}, 34:21150--21163, 2021.

\bibitem{dorfman2023docofl}
Ron Dorfman, Shay Vargaftik, Yaniv Ben-Itzhak, and Kfir~Y. Levy.
\newblock Docofl: downlink compression for cross-device federated learning.
\newblock In {\em Proceedings of the 40th International Conference on Machine Learning}, ICML'23. JMLR.org, 2023.

\bibitem{bernstein2018signsgd}
Jeremy Bernstein, Yu-Xiang Wang, Kamyar Azizzadenesheli, and Animashree Anandkumar.
\newblock signsgd: Compressed optimisation for non-convex problems.
\newblock In {\em International Conference on Machine Learning}, pages 560--569. PMLR, 2018.

\bibitem{li2021talk}
Liang Li, Dian Shi, Ronghui Hou, Hui Li, Miao Pan, and Zhu Han.
\newblock To talk or to work: Flexible communication compression for energy efficient federated learning over heterogeneous mobile edge devices.
\newblock In {\em IEEE INFOCOM 2021-IEEE Conference on Computer Communications}, pages 1--10. IEEE, 2021.

\bibitem{xu2022adaptive}
Yang Xu, Yunming Liao, Hongli Xu, Zhenguo Ma, Lun Wang, and Jianchun Liu.
\newblock Adaptive control of local updating and model compression for efficient federated learning.
\newblock {\em IEEE Transactions on Mobile Computing}, 2022.

\bibitem{mei2022resource}
Yiqun Mei, Pengfei Guo, Mo~Zhou, and Vishal Patel.
\newblock Resource-adaptive federated learning with all-in-one neural composition.
\newblock {\em Advances in Neural Information Processing Systems}, 35:4270--4284, 2022.

\bibitem{liu2023communication}
Heting Liu, Fang He, and Guohong Cao.
\newblock Communication-efficient federated learning for heterogeneous edge devices based on adaptive gradient quantization.
\newblock In {\em IEEE INFOCOM 2023-IEEE Conference on Computer Communications}, pages 1--10. IEEE, 2023.

\bibitem{yan2024peaches}
Jiaming Yan, Jianchun Liu, Hongli Xu, Zhiyuan Wang, and Chunming Qiao.
\newblock Peaches: Personalized federated learning with neural architecture search in edge computing.
\newblock {\em IEEE Transactions on Mobile Computing}, 2024.

\bibitem{ye2023feddisco}
Rui Ye, Mingkai Xu, Jianyu Wang, Chenxin Xu, Siheng Chen, and Yanfeng Wang.
\newblock Feddisco: Federated learning with discrepancy-aware collaboration.
\newblock {\em arXiv preprint arXiv:2305.19229}, 2023.

\bibitem{wang2023bose}
Lun Wang, Yang Xu, Hongli Xu, Zhida Jiang, Min Chen, Wuyang Zhang, and Chen Qian.
\newblock Bose: Block-wise federated learning in heterogeneous edge computing.
\newblock {\em IEEE/ACM Transactions on Networking}, 2023.

\bibitem{li2014efficient}
Mu~Li, Tong Zhang, Yuqiang Chen, and Alexander~J Smola.
\newblock Efficient mini-batch training for stochastic optimization.
\newblock In {\em Proceedings of the 20th ACM SIGKDD international conference on Knowledge discovery and data mining}, pages 661--670, 2014.

\bibitem{hsu2019measuring}
Tzu-Ming~Harry Hsu, Hang Qi, and Matthew Brown.
\newblock Measuring the effects of non-identical data distribution for federated visual classification.
\newblock {\em arXiv preprint arXiv:1909.06335}, 2019.

\bibitem{yurochkin2019bayesian}
Mikhail Yurochkin, Mayank Agarwal, Soumya Ghosh, Kristjan Greenewald, Nghia Hoang, and Yasaman Khazaeni.
\newblock Bayesian nonparametric federated learning of neural networks.
\newblock In {\em International conference on machine learning}, pages 7252--7261. PMLR, 2019.

\bibitem{he2016deep}
Kaiming He, Xiangyu Zhang, Shaoqing Ren, and Jian Sun.
\newblock Deep residual learning for image recognition.
\newblock In {\em Proceedings of the IEEE conference on computer vision and pattern recognition}, pages 770--778, 2016.

\bibitem{li2022pyramidfl}
Chenning Li, Xiao Zeng, Mi~Zhang, and Zhichao Cao.
\newblock Pyramidfl: A fine-grained client selection framework for efficient federated learning.
\newblock In {\em Proceedings of the 28th Annual International Conference on Mobile Computing And Networking}, pages 158--171, 2022.

\bibitem{lai2021oort}
Fan Lai, Xiangfeng Zhu, Harsha~V Madhyastha, and Mosharaf Chowdhury.
\newblock Oort: Efficient federated learning via guided participant selection.
\newblock In {\em 15th $\{$USENIX$\}$ Symposium on Operating Systems Design and Implementation ($\{$OSDI$\}$ 21)}, pages 19--35, 2021.

\bibitem{alistarh2018convergence}
Dan Alistarh, Torsten Hoefler, Mikael Johansson, Nikola Konstantinov, Sarit Khirirat, and C{\'e}dric Renggli.
\newblock The convergence of sparsified gradient methods.
\newblock {\em Advances in Neural Information Processing Systems}, 31, 2018.

\bibitem{liao2024mergesfl}
Yunming Liao, Yang Xu, Hongli Xu, Lun Wang, Zhiwei Yao, and Chunming Qiao.
\newblock Mergesfl: Split federated learning with feature merging and batch size regulation.
\newblock In {\em 2024 IEEE 40th International Conference on Data Engineering (ICDE)}, pages 2054--2067. IEEE, 2024.

\bibitem{hershey2007approximating}
John~R Hershey and Peder~A Olsen.
\newblock Approximating the kullback leibler divergence between gaussian mixture models.
\newblock In {\em 2007 IEEE International Conference on Acoustics, Speech and Signal Processing-ICASSP'07}, volume~4, pages IV--317. IEEE, 2007.

\bibitem{goldberger2003efficient}
Goldberger and Greenspan.
\newblock An efficient image similarity measure based on approximations of kl-divergence between two gaussian mixtures.
\newblock In {\em Proceedings Ninth IEEE International conference on computer vision}, pages 487--493. IEEE, 2003.

\bibitem{ma2023adaptive}
Zhenguo Ma, Yang Xu, Hongli Xu, Zeyu Meng, Liusheng Huang, and Yinxing Xue.
\newblock Adaptive batch size for federated learning in resource-constrained edge computing.
\newblock {\em IEEE Transactions on Mobile Computing}, 22(1):37--53, 2023.

\bibitem{paszke2019pytorch}
Adam Paszke, Sam Gross, Francisco Massa, Adam Lerer, James Bradbury, Gregory Chanan, Trevor Killeen, Zeming Lin, Natalia Gimelshein, Luca Antiga, et~al.
\newblock Pytorch: An imperative style, high-performance deep learning library.
\newblock {\em Advances in neural information processing systems}, 32, 2019.

\bibitem{jiang2020mnn}
Xiaotang Jiang, Huan Wang, Yiliu Chen, Ziqi Wu, Lichuan Wang, Bin Zou, Yafeng Yang, Zongyang Cui, Yu~Cai, Tianhang Yu, et~al.
\newblock Mnn: A universal and efficient inference engine.
\newblock {\em Proceedings of Machine Learning and Systems}, 2:1--13, 2020.

\bibitem{naik2016building}
Nitin Naik.
\newblock Building a virtual system of systems using docker swarm in multiple clouds.
\newblock In {\em 2016 IEEE International Symposium on Systems Engineering (ISSE)}, pages 1--3. IEEE, 2016.

\bibitem{dalcin2021mpi4py}
Lisandro Dalcin and Yao-Lung~L Fang.
\newblock mpi4py: Status update after 12 years of development.
\newblock {\em Computing in Science \& Engineering}, 23(4):47--54, 2021.

\bibitem{harris2020array}
Charles~R Harris, K~Jarrod Millman, St{\'e}fan~J Van Der~Walt, Ralf Gommers, Pauli Virtanen, David Cournapeau, Eric Wieser, Julian Taylor, Sebastian Berg, Nathaniel~J Smith, et~al.
\newblock Array programming with numpy.
\newblock {\em Nature}, 585(7825):357--362, 2020.

\bibitem{krizhevsky2009learning}
Alex Krizhevsky, Geoffrey Hinton, et~al.
\newblock Learning multiple layers of features from tiny images.
\newblock 2009.

\bibitem{anguita2013public}
Davide Anguita, Alessandro Ghio, Luca Oneto, Xavier Parra, Jorge~Luis Reyes-Ortiz, et~al.
\newblock A public domain dataset for human activity recognition using smartphones.
\newblock In {\em Esann}, volume~3, page~3, 2013.

\bibitem{warden2018speech}
Pete Warden.
\newblock Speech commands: A dataset for limited-vocabulary speech recognition.
\newblock {\em arXiv preprint arXiv:1804.03209}, 2018.

\bibitem{yoon2022bitwidth}
Jaehong Yoon, Geon Park, Wonyong Jeong, and Sung~Ju Hwang.
\newblock Bitwidth heterogeneous federated learning with progressive weight dequantization.
\newblock In {\em International Conference on Machine Learning}, pages 25552--25565. PMLR, 2022.

\bibitem{zhao2023towards}
Zihao Zhao, Yuzhu Mao, Yang Liu, Linqi Song, Ye~Ouyang, Xinlei Chen, and Wenbo Ding.
\newblock Towards efficient communications in federated learning: A contemporary survey.
\newblock {\em Journal of the Franklin Institute}, 360(12):8669--8703, 2023.

\bibitem{honig2022dadaquant}
Robert H{\"o}nig, Yiren Zhao, and Robert Mullins.
\newblock Dadaquant: Doubly-adaptive quantization for communication-efficient federated learning.
\newblock In {\em International Conference on Machine Learning}, pages 8852--8866. PMLR, 2022.

\bibitem{alistarh2017qsgd}
Dan Alistarh, Demjan Grubic, Jerry Li, Ryota Tomioka, and Milan Vojnovic.
\newblock Qsgd: Communication-efficient sgd via gradient quantization and encoding.
\newblock {\em Advances in neural information processing systems}, 30, 2017.

\bibitem{sattler2019robust}
Felix Sattler, Simon Wiedemann, Klaus-Robert M{\"u}ller, and Wojciech Samek.
\newblock Robust and communication-efficient federated learning from non-iid data.
\newblock {\em IEEE transactions on neural networks and learning systems}, 31(9):3400--3413, 2019.

\bibitem{horvath2021fjord}
Samuel Horvath, Stefanos Laskaridis, Mario Almeida, Ilias Leontiadis, Stylianos Venieris, and Nicholas Lane.
\newblock Fjord: Fair and accurate federated learning under heterogeneous targets with ordered dropout.
\newblock {\em Advances in Neural Information Processing Systems}, 34:12876--12889, 2021.

\bibitem{diao2020heterofl}
Enmao Diao, Jie Ding, and Vahid Tarokh.
\newblock Heterofl: Computation and communication efficient federated learning for heterogeneous clients.
\newblock In {\em International Conference on Learning Representations}, 2020.

\bibitem{wang2023svdfed}
Haolin Wang, Xuefeng Liu, Jianwei Niu, and Shaojie Tang.
\newblock Svdfed: Enabling communication-efficient federated learning via singular-value-decomposition.
\newblock In {\em IEEE INFOCOM 2023-IEEE Conference on Computer Communications}, pages 1--10. IEEE, 2023.

\bibitem{hyeon2021fedpara}
Nam Hyeon-Woo, Moon Ye-Bin, and Tae-Hyun Oh.
\newblock Fedpara: Low-rank hadamard product for communication-efficient federated learning.
\newblock In {\em International Conference on Learning Representations}, 2021.

\bibitem{li2021hermes}
Ang Li, Jingwei Sun, Pengcheng Li, Yu~Pu, Hai Li, and Yiran Chen.
\newblock Hermes: an efficient federated learning framework for heterogeneous mobile clients.
\newblock In {\em Proceedings of the 27th Annual International Conference on Mobile Computing and Networking}, pages 420--437, 2021.

\bibitem{wang2018edge}
Shiqiang Wang, Tiffany Tuor, Theodoros Salonidis, Kin~K Leung, Christian Makaya, Ting He, and Kevin Chan.
\newblock When edge meets learning: Adaptive control for resource-constrained distributed machine learning.
\newblock In {\em IEEE INFOCOM 2018-IEEE conference on computer communications}, pages 63--71. IEEE, 2018.

\bibitem{wang2020tackling}
Jianyu Wang, Qinghua Liu, Hao Liang, Gauri Joshi, and H~Vincent Poor.
\newblock Tackling the objective inconsistency problem in heterogeneous federated optimization.
\newblock {\em Advances in neural information processing systems}, 33:7611--7623, 2020.

\bibitem{luo2022tackling}
Bing Luo, Wenli Xiao, Shiqiang Wang, Jianwei Huang, and Leandros Tassiulas.
\newblock Tackling system and statistical heterogeneity for federated learning with adaptive client sampling.
\newblock In {\em IEEE INFOCOM 2022-IEEE conference on computer communications}, pages 1739--1748. IEEE, 2022.

\end{thebibliography}

\appendix

\end{document}